%% file: example.tex
\documentclass{article}

\usepackage[final]{corl_2026} % Uncomment for the camera-ready ``final'' version.
\usepackage{silence}
\usepackage[table]{xcolor}
\definecolor{lightgray}{gray}{0.85}
\usepackage{graphicx}
\usepackage{booktabs}
\usepackage{wrapfig}
\usepackage{multirow}
\usepackage{float}
\usepackage{subcaption}
\usepackage[export]{adjustbox}
\usepackage{colortbl}
\usepackage{hyperref}
\usepackage{amsmath}
\usepackage{amssymb}
\usepackage{enumitem}
\usepackage{algorithm,algorithmic}
\usepackage{tcolorbox}
\usepackage{underscore}
\tcbuselibrary{listings}
\tcbuselibrary{breakable}
\tcbset{
    listing engine=listings,
    colback=white!95!blue,% 背景颜色
    colframe=blue!75!black,% 边框颜色
    listing options={% 列出选项
        basicstyle=\ttfamily,
        numbers=left,
        numberstyle=\tiny,
        stepnumber=1,
        showstringspaces=false,
        breaklines=true,
        breakatwhitespace=true,
        tabsize=4,
        language=C++
    }
}

\title{EMMA: Enhancing Robotic Control Robustness through a Generative Visual Model}

\author{
  Zhehao Dong$^{1,2*}$,
  Xiaofeng Wang$^{2,3*}$,
  Zheng Zhu$^{2*\dag}$,
  Yirui Wang$^{3}$,
  Yang Wang$^{2}$ \\
  \textbf{Yukun Zhou}$^{2}$,
  \textbf{Boyuan Wan}g$^{2,4}$,
  \textbf{Chaojun Ni}$^{1,2}$,
  \textbf{Runqi Ouyang}$^{2,4}$,
  \textbf{Wenkang Qin}$^{2}$ \\
  \textbf{Xinze Chen}$^{2}$,
  \textbf{Yun Ye}$^{2}$,
  \textbf{Guan Huang}$^{2}$,
  \textbf{Zhen Lu}$^{1}$,
  \textbf{Yue Yang}$^{1}$ \\
  $^{1}$School of Mechanics and Engineering Science, Peking University,
  $^{2}$GigaAI \\
  $^{3}$Tsinghua University,
  $^{4}$CASIA \\
  {\tt \{zhengzhu\}@ieee.org} \\
  $^{*}$ Equal contribution, $^{\dag}$ Corresponding author. \\
  Project page: \url{https://emma-gigaai.github.io/}
}

\begin{document}
\maketitle

\input{sec/0-abstract}
\input{sec/1-intro}
\input{sec/3-method}

\input{sec/4-experiments}
\input{sec/5-conclusion}

\clearpage
% The acknowledgments are automatically included only in the final and preprint versions of the paper.

%===============================================================================

% no \bibliographystyle is required, since the corl style is automatically used.
\bibliography{example}  % .bib

\newpage
\input{sec/6-appendix}

\end{document}

%% file: sec/0-abstract.tex
\vspace{-1em}
\begin{figure*}[ht]
    \centering
    \includegraphics[width=\textwidth]{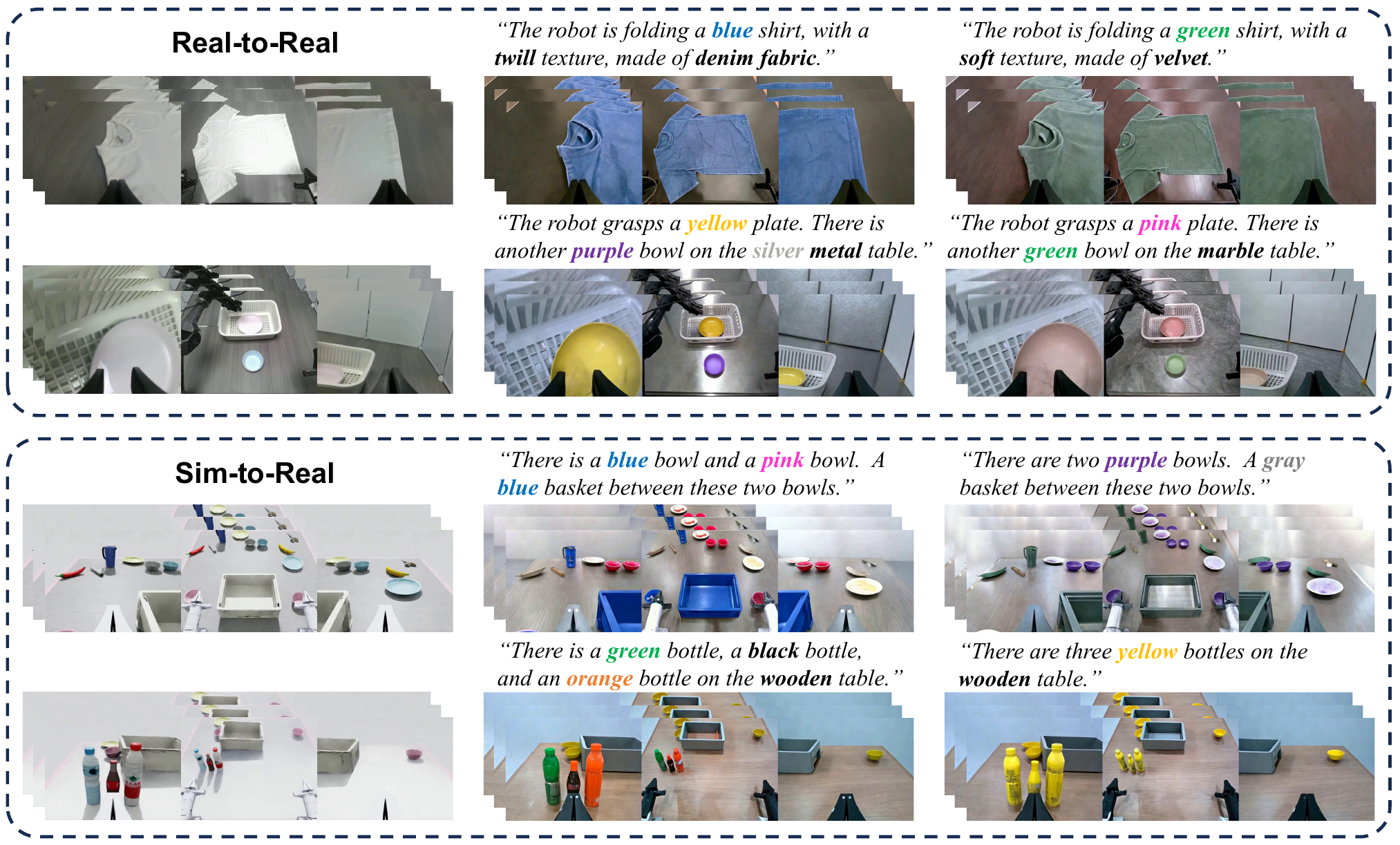}
    \vspace{-1em}
    \caption{
    \textit{DreamTransfer} demonstrates strong controllability in embodied manipulation video generation.
    It excels in text-guided appearance editing while preserving 3D structure and geometric plausibility, and supports both real-to-real and sim-to-real transfer.
    The first column: real-world and simulation data. Next two columns: diverse data generated by \textit{DreamTransfer} from the first column.}
    \label{fig:transfer-vis}
\end{figure*}

\begin{abstract}
The generalization of vision-language-action (VLA) models heavily relies on diverse training data. However, acquiring large-scale robot manipulation data across varied object appearances is costly and labor-intensive. To address this, we introduce \textbf{E}mbodied \textbf{M}anipulation \textbf{M}edia \textbf{A}daptation (\textit{EMMA}), a framework for augmenting VLA policies that combines a generative data engine with an effective training pipeline. We introduce \textit{DreamTransfer}, a diffusion Transformer-based architecture for generating multi-view consistent and geometrically plausible embodied manipulation videos. It supports text-driven video editing on foreground, background and lighting while preserving 3D structure and geometric fidelity. We also utilize a hybrid training set of real and generated data and propose \textit{AdaMix} to enhance the training process. \textit{AdaMix} is a training strategy that adaptively weights samples according to policy performance to emphasize challenging samples. 
Evaluations show \textit{DreamTransfer} outperforms previous video generation models in multi-view consistency, geometric fidelity, and text-conditioning precision.
In real-world robotic tasks with zero-shot visual settings, our framework achieves a relative performance increase of over 92\% compared to training with real data alone, and improves by an additional 17\% with \textit{AdaMix}.
\end{abstract}

\keywords{Diffusion Transformer, Embodied Manipulation, VLA} 

%% file: sec/1-intro.tex
\section{Introduction}
\label{sec:intro}

Vision-language-action (VLA) models~\cite{black2024pi0,intelligence2025pi05,brohan2023rt2,kim2024openvla,nvidia2025gr00tn1,deng2025graspvla,yue2024deervla,wei2024occllama,zhang2025safevla,huang2025early,li2025selfcorrecting,wang2026nativemem,jiang2025rynnvla001} have demonstrated remarkable capabilities in enabling robots to perform complex manipulation tasks from natural language instructions and visual inputs~\cite{shridhar2021cliport,zhen2025tesseract}. However, their success critically depends on large-scale, diverse training data. 
Collecting real-world robot manipulation data through human teleoperation is labor-intensive and expensive, severely limiting the scale and visual diversity of available datasets~\cite{embodimentcollaboration2025openx,khazatsky2025droid}. 
While simulation offers a scalable alternative for generating annotated trajectories~\cite{geng2025roboverse,mu2025robotwin,katara2023gen2sim,lin2024ubsoft}, simulation environments often suffer from visual realism gaps and are constrained by limited asset diversity. As a result, VLA policies trained on simulation data frequently underperform when deployed in the real world.

Various generative approaches have been proposed for low-cost diverse robot data generation~\cite{lin2025data,chen2024roviaugrobot,jin2025physically,yuan2025motiontranshumanvrdata,eskandar2026videoweaver,wang2025embodiedreamer,teoh2024greenscreen,chen2024semantically,mandi2023cacti,wang2024cyberdemo,zhuang2024enhancing,yu2023scalingrobotlearning,yang2025noveldemonstrationgeneration,ni2026gigaworld,ni2025recondreamer,fang2025rebotscalingrobotlearning}. Recently, diffusion models~\cite{ho2020denoising,song2022denoising,ho2022classifierfree} have shown promise for generating realistic videos~\cite{blattmann2023stablevideodiffusion,wan2025wan,kong2025hunyuanvideo,nvidia2025cosmos,yang2025cogvideox,zheng2024opensora,lu2023vdtgeneral,yuan2025roboengine}. Several works have explored using diffusion models to generate vision-action data. Cosmos-Transfer1~\cite{nvidia2025cosmostransfer} uses semantic segmentation and depth maps to generate realistic scenes. RoboTransfer~\cite{liu2025robotransfer} further improves multi-view consistency by explicitly modeling 3D geometry.
We discuss related work in Appendix~\ref{app:related-work} and highlight core research gaps.
Despite these advances, two key challenges remain. First, most methods~\cite{yuan2025roboengine,nvidia2025cosmostransfer} fail to ensure multi-view consistency, limiting their application in multi-camera robotic tasks. While RoboTransfer explores this, it relies on image conditioning and generalizes poorly.
Moreover, existing multi-view diffusion models~\cite{shi2024mvdream,kant2024spad,liu2023zero} target static 3D modeling and lack temporal modeling for dynamic robot manipulation.
Second, existing works treat generated data as a static augmentation, without considering how to use it effectively during training.

In this work, we propose \textbf{E}mbodied \textbf{M}anipulation \textbf{M}edia \textbf{A}daptation (\textit{EMMA}), a VLA policy enhancement framework that integrates two core components: \textit{DreamTransfer} and \textit{AdaMix}. 
\textit{DreamTransfer} is a diffusion Transformer (DiT)-based framework for generating multi-view consistent, geometrically plausible embodied manipulation videos. 
It jointly models appearance and geometry across multiple camera views, ensuring spatial and temporal coherence. \textit{DreamTransfer} supports text-guided visual transfer: users can edit the foreground objects, background, and lighting conditions of real or simulation demonstrations through natural language, while preserving the underlying 3D structure and geometric plausibility of the scene.
As illustrated in Figure~\ref{fig:transfer-vis}, \textit{DreamTransfer} enables realistic and controllable video generation for both real-to-real and sim-to-real transfer scenarios, making it a powerful tool for scalable policy training.
To improve policy learning, we further propose \textit{AdaMix}, a hard-sample-aware training strategy. We propose three metrics to evaluate the quality of predicted trajectories from the VLA policy and use the performance score to drive an adaptive sampling mechanism. By iteratively refining the training distribution toward challenging cases, our method improves robustness and generalization.

% We evaluate on diverse robotic manipulation tasks across video generation quality and over 1,800 trials in both simulation and real environments. 
We evaluate \textit{EMMA} in terms of video generation quality and policy performance across diverse robotic manipulation tasks, with over 1,800 trials in total across simulated and real-world environments.
% These tasks cover rigid/deformable objects, short/long-horizon actions, and skills like grasping, pushing, placing, and draping.
Compared to the state-of-the-art transfer model, \textit{DreamTransfer} improves multi-view consistency by 42\% and depth consistency by 24\%, demonstrating superior cross-view coherence and geometric fidelity. 
For real-world tasks with \textbf{zero-shot} visual appearances, our method achieves over a 92\% relative improvement in task success rate compared to training on real data alone, with an additional 17\% gain when integrated with \textit{AdaMix}.

The main contributions of this paper are summarized as follows:
\begin{itemize}[leftmargin=*, labelsep=5pt, itemindent=0pt]
\item We propose \textit{EMMA}, a VLA policy enhancement framework that integrates a generative data engine with an effective training strategy. 
The data engine generates diverse, multi-view consistent robot manipulation videos for both rigid and deformable objects, while adaptive sample weighting improves VLA policy generalization. 
\item We propose \textit{DreamTransfer}, a DiT-based model that generates multi-view consistent, geometrically plausible manipulation videos and supports text-guided editing of foreground, background, and lighting. 
We further introduce \textit{AdaMix}, a hard-sample-aware training strategy that identifies and adaptively reweights challenging trajectories during training. 
\item \textit{EMMA} demonstrates strong performance in video generation and real-world robotic deployment. Compared to the state-of-the-art model, \textit{DreamTransfer} achieves a 42\% gain in multi-view consistency and a 24\% gain in depth consistency, measured relatively. In zero-shot visual settings, our method achieves over a 92\% performance gain compared to real-data training, with \textit{AdaMix} providing an additional 17\% improvement and enhancing cross-domain visual generalization.
\end{itemize}

%% file: sec/3-method.tex
\section{Method}
\label{method}

\subsection{Framework Overview}

\begin{figure*}[ht]
    \centering
    \includegraphics[width=\textwidth]{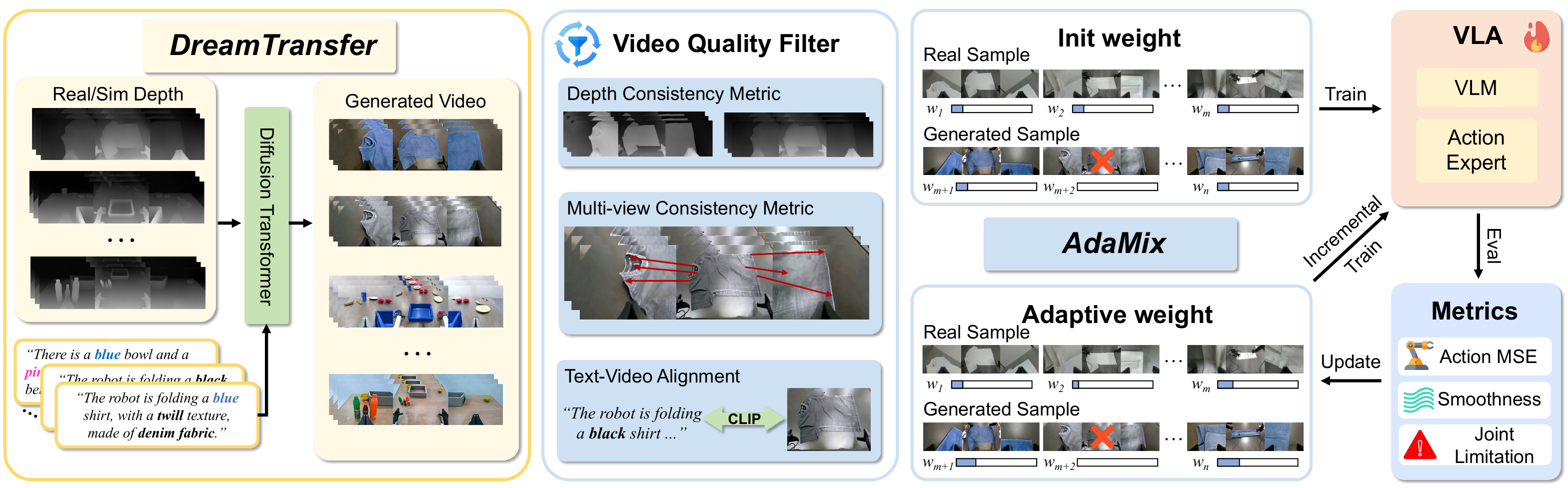}
    \caption{Overview of the \textit{EMMA} framework. 
    First, \textit{DreamTransfer} generates multi-view consistent videos via text-guided visual editing of the foreground, background, and lighting conditions, conditioned on depth and text prompts.
    The generated videos are then evaluated by a video quality filter. Low-quality videos are initially assigned zero sampling weight to stabilize early-stage training.
    The \textit{AdaMix} module further adaptively reweights training samples based on trajectory quality metrics, up-weighting challenging samples to improve policy robustness and generalization.}
    \label{fig:emma-framework}
\end{figure*}

We propose \textit{EMMA}, a VLA policy enhancement framework that advances embodied manipulation by integrating two core components: \textit{DreamTransfer} for multi-view video generation and \textit{AdaMix} for hard-sample-aware training.
As illustrated in Figure~\ref{fig:emma-framework}, \textit{DreamTransfer} functions as a generative data engine for training VLA, capable of generating multi-view consistent, geometrically plausible robot manipulation videos from both real and simulation demonstration videos. 
Its key novelty lies in enabling fine-grained, text-guided editing of visual attributes such as foreground objects, backgrounds, and lighting. Crucially, it achieves this while strictly preserving multi-view consistency.
The generated videos are filtered according to depth consistency, multi-view consistency, and text-video similarity metrics. Specifically, only high-quality videos are used for initial training, with low-quality samples assigned zero sampling probability.
To further enhance policy generalization, we introduce \textit{AdaMix}, a hard-sample-aware training strategy that dynamically adjusts the sampling weights based on their trajectory performance metric. By adaptively emphasizing challenging cases, 
\textit{AdaMix} mitigates long-tail errors and improves robustness.

We first introduce the \textit{DreamTransfer} model in Section~\ref{sec:dreamtransfer}, followed by a presentation of the \textit{AdaMix} training strategy in Section~\ref{sec:adamix}.

\subsection{DreamTransfer}
\label{sec:dreamtransfer}

\begin{wrapfigure}{r}{6cm}
    \centering 
    \includegraphics[width=\linewidth]{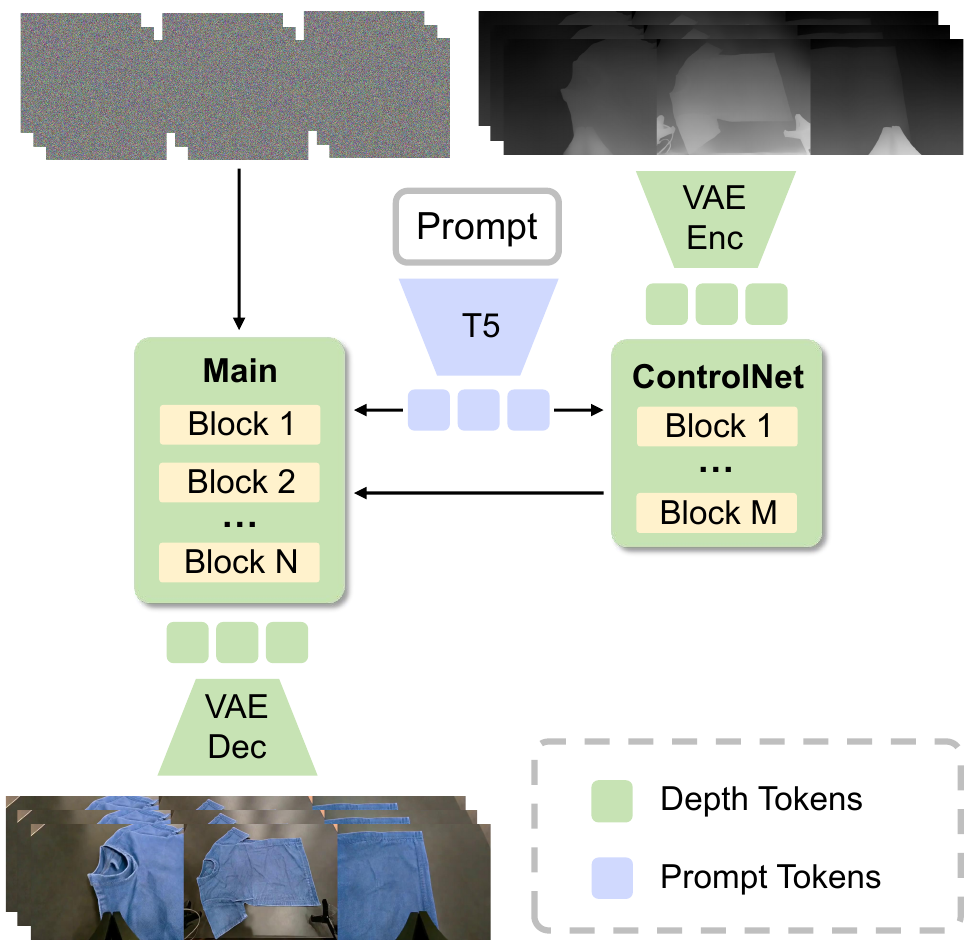}
    \caption{Overview of the \textit{DreamTransfer} framework. Multi-view depth maps are concatenated along the width dimension.}
    \label{fig:generation-model-framework}
\end{wrapfigure}

The overall framework of \textit{DreamTransfer} is illustrated in Figure~\ref{fig:generation-model-framework}. 
\textit{DreamTransfer} features a dual-branch architecture, both built upon DiT-based framework~\cite{nvidia2025cosmos,nvidia2025cosmostransfer}. The main branch denoises latent video tokens, while the parallel ControlNet~\cite{zhang2023adding} branch enhances geometric consistency by incorporating depth-based structural guidance. 

\textit{DreamTransfer} leverages multi-view in-context learning capabilities~\cite{jang2025dreamgen,huang2024incontextlora,zhao2024drivedreamer2,liu2025robotransfer} inherently present in pretrained diffusion models, ensuring spatial and temporal coherence across different camera viewpoints. First, synchronized depth videos from multiple viewpoints $\{v^{(1)}, v^{(2)}, \ldots, v^{(M)}\}$ are concatenated along the width dimension and encoded into a unified latent representation using a VAE encoder~\cite{kingma2022autoencoding}:
\begin{equation}
    d_{init} = \mathrm{Enc}([v^{(1)}, v^{(2)}, \ldots, v^{(M)}]).
    \label{eq:eq-1}
\end{equation}
In parallel, a pretrained T5 text encoder~\cite{raffel2023exploring} converts text prompt into high-dimensional semantic embeddings. These textual features are fused with the visual latents via cross-attention, enabling fine-grained control over object appearance. The main branch then takes the noisy latent $z_t$ at time step $t$ together with the prompt features $s$ and depth features $d_t$, to predict the denoised latent:
\begin{equation}
    {n} = f_\theta(z_t, t, d_t, s),
    \label{eq:eq-2}
\end{equation}
where $f_\theta$ denotes the diffusion Transformer parameterized by $\theta$, and $n$ is the predicted denoised latent.
Finally, the denoised latent is decoded by the VAE decoder to reconstruct the output video, preserving the underlying 3D structure and faithfully reflects the prompt-specified appearance changes.

We construct a dataset of 50k multi-view video clips from Agibot World dataset~\cite{bu2025agibot} with each clip containing 121 frames. 
\textit{DreamTransfer} uses two-stage fine-tuning that enables high-quality multi-view video generation with just 131 hours of training on four NVIDIA H20 GPUs.
\textit{DreamTransfer} generalizes to unseen datasets, including different robotic arms, environments and tasks.
Details of training and dataset are given in Appendix~\ref{app:dreamtransfer-train-detail}.
\textit{DreamTransfer} generates 121-frame videos, extendable to long horizons via clip-wise generation and concatenation. For VLA policy training, each frame acts as an independent observation-action pair, providing sufficient data for effective learning.

\subsection{AdaMix}
\label{sec:adamix}

How to use embodied manipulation data effectively for training VLA policies remains an open challenge. Related ideas have been explored in other fields. For instance, Delphi~\cite{ma2024unleashing} in autonomous driving improves generalization by focusing on failure cases, and POLARIS~\cite{Polaris2025} in LLM-based reinforcement learning enhances training by filtering out overly simple questions. Inspired by these approaches, our work explores efficient training strategies tailored to downstream VLA policies.

We observe that even with extensive fine-tuning on mixed real and generated data, VLA policies still struggle with long-tail scenarios under uniform training paradigms, even if these scenarios are present in the training set.
Building on this insight, we present \textit{AdaMix}, a hard-sample-aware training strategy that dynamically reweights training data based on policy performance. Unlike uniform sampling schemes that treat all samples equally, \textit{AdaMix} continuously identifies challenging scenarios, particularly those in the long-tail where policies typically struggle. By up-weighting these hard samples during training, \textit{AdaMix} achieves improved generalization.

The \textit{AdaMix} pipeline consists of three stages, as illustrated in right part of Figure~\ref{fig:emma-framework}.
First, generated videos are evaluated by a video quality filter based on depth consistency, multi-view consistency, and text-video similarity. Samples failing to meet quality thresholds are assigned zero sampling weight, ensuring that only geometrically plausible and semantically faithful data are used for training.
Then, the VLA policy is initially trained using uniform sampling over real data and high-quality generated data. 
After the loss converges, we compute the following three predicted trajectory quality metrics on each training sample to identify challenging instances. These metrics are inspired by NVIDIA’s robotic policy benchmarking framework~\cite{yang2025robot} and tailored to our setting.

\textbf{Action MSE.} 
We compute the Mean Squared Error (MSE) between the predicted action chunk $\hat{a}_{i:t}$ and the corresponding ground-truth sequence $a_{i:t}$ over a window of at most $L$ frames:
\begin{equation}
    r^{\text{MSE}}_i = - \frac{1}{L} \sum_{t=0}^{L-1} \|\hat{a}_{i+t} - a_{i+t}\|_2^2.
    \label{eq:eq-3}
\end{equation}

\textbf{Smoothness.} 
To encourage physically plausible actions, we measure the second-order difference of joint angles to penalize abrupt joint actions:
\begin{equation}
    r^{\text{Smooth}}_i = - \sum_{j} \left| \frac{a_{i+2,j} - 2a_{i+1,j} + a_{i,j}}{(\Delta t)^2} \right|,
    \label{eq:eq-4}
\end{equation}
where $j$ indexes the joint dimension, and $\Delta t$ represents the time interval between frames.

\textbf{Joint limitation.} 
We assign a binary indicator to ensure safety, setting
\begin{equation}
    r^{\text{Limit}}_i =
    \begin{cases}
        1, & \text{if all joint angles within thresholds}, \\
        0, & \text{otherwise}.
    \end{cases}
    \label{eq:eq-5}
\end{equation}

These scores are min-max normalized to $\tilde{r}^{(\cdot)}_i \in [0,1]$ and combined into a unified score per sample \(\):
\begin{equation}
    s_i = w_{\text{MSE}} \cdot \tilde{r}^{\text{MSE}}_i + 
      w_{\text{Smooth}} \cdot \tilde{r}^{\text{Smooth}}_i +
      w_{\text{Limit}} \cdot \tilde{r}^{\text{Limit}}_i,
    \label{eq:eq-6}
\end{equation}
where $w_{\text{MSE}} + w_{\text{Smooth}} + w_{\text{Limit}} = 1$ and each weight is non-negative. 
A higher \(s_i\)  indicates better overall performance on the $i$-th sample.
To find a suitable weighting scheme, we conducted an ablation study on the three metrics in Appendix~\ref{app:ablation_metrics}.
We find that these metrics exhibit comparable impact on performance.
Therefore, we adopt equal weights for all three metrics.

During incremental training, sampling weights are updated as:
\begin{equation}
p(i) \propto \gamma + \lambda \cdot (1 - s_i),
    \label{eq:eq-7}
\end{equation}
where \(\gamma > 0\) ensures minimum support for all samples, and \(\lambda\) controls the emphasis on hard samples.
We perform a quantitative analysis of the two key hyperparameters $\gamma$ and $\lambda$ in Appendix~\ref{app:study_hyperpara}.
The configuration $\gamma = 1$, $\lambda = 10$ achieves the best trade-off, and is adopted in this work.

By adaptively up-weighting samples where the policy performs poorly, the training distribution gradually shifts toward underperforming regions while preserving data diversity. 
All evaluations and re-weighting are performed on training data to avoid validation leakage.

%% file: sec/4-experiments.tex
\section{Experiments}
\label{experiments}

In this section, we evaluate \textit{EMMA} framework on both video generation (Section~\ref{sec:video-generation-quality}) and real-world robotic deployment (Section~\ref{sec:real-robot-eval}). We conducted extensive experiments on diverse tasks such as \texttt{Fold Clothes}~(real-to-real), \texttt{Clean Desk}~(sim-to-real), and \texttt{Throw Bottle}~(sim-to-real). 
These tasks cover a range of robotic tasks involving both rigid and deformable objects, long-horizon and short-horizon actions, and different manipulation skills such as grasping, pushing, and placing.
In addition, we conduct simulation experiments in Section~\ref{sec:simulation-experiments}.

\subsection{Video Generation Quality}
\label{sec:video-generation-quality}

\begin{table*}[h]
\caption{Comparison of video generation quality on robot manipulation tasks. Best results are in \textbf{bold}. The result demonstrates that \textit{DreamTransfer} achieves the best average performance across all evaluation metrics.}
\label{video-transfer-table}
\begin{center}
\resizebox{\linewidth}{!}{%
\begin{tabular}{llccccc}
\toprule
\multicolumn{1}{l}{Task}      & Model             & Mat.Pix.(↑) & RMSE(↓)  & Abs.Rel.(↓) & Sq.Rel.(↓) & CLIPSim.(↑) \\ \midrule
\multirow{3}{*}{Fold Clothes}   & RoboTransfer~\cite{liu2025robotransfer}       & 1736     & 3.97 & 0.50  &    2.13  &  24.63    \\
& Cosmos-Transfer1~\cite{nvidia2025cosmostransfer}     & 2210     & 2.78 & 0.36    & 1.14   & 25.02    \\
& \textit{DreamTransfer}    & 2604     & 2.50 & 0.32     & 0.97   & 24.54  \\
\midrule
\multirow{3}{*}{Clean Desk}   & RoboTransfer~\cite{liu2025robotransfer}       & 3213     & 2.50 & 0.32  &  0.85   & 23.87    \\
& Cosmos-Transfer1~\cite{nvidia2025cosmostransfer}     & 2484     & 1.55 & 0.19    & 0.36   & 25.02 \\
& \textit{DreamTransfer}    & 4311     & 1.45 & 0.18    & 0.33   & 25.75 \\ 
\midrule
\multirow{3}{*}{Throw Bottle} & RoboTransfer~\cite{liu2025robotransfer}       & 1944     & 2.08 & 0.34  &  0.93   & 23.31    \\
& Cosmos-Transfer1~\cite{nvidia2025cosmostransfer}     & 1597     & 1.71 & 0.26    & 0.64   & 23.79    \\
& \textit{DreamTransfer}    & 2894     & 1.36 & 0.20    & 0.33   & 23.74 \\ 
\midrule
\multirow{3}{*}{Average}      & RoboTransfer~\cite{liu2025robotransfer}       & 2298     & 2.85 & 0.39  &  1.30   & 23.94    \\
& Cosmos-Transfer1~\cite{nvidia2025cosmostransfer}     & 2097     & 2.01 & 0.27    & 0.71   & 24.61    \\
& \cellcolor{lightgray}\textit{DreamTransfer} &  \cellcolor{lightgray}\textbf{3270} &  \cellcolor{lightgray}\textbf{1.77} &  \cellcolor{lightgray}\textbf{0.23} &  \cellcolor{lightgray}\textbf{0.54} &  \cellcolor{lightgray}\textbf{24.68} \\ 
\bottomrule
\end{tabular}
}
\end{center}
\vspace{-1em}
\end{table*}

\noindent
\textbf{Baselines.}
While general video generation models~\cite{yang2025cogvideox,wan2025wan,blattmann2023stablevideodiffusion,zheng2024opensora} exhibit strong open-domain performance, they suffer from several limitations in embodied scenarios, specifically multi-view inconsistency and poor physical fidelity. We provide visualization results of general video generation models in Appendix~\ref{app:video-generation-models-comparison}. Robotic task-specialized models such as Cosmos-Transfer1~\cite{nvidia2025cosmostransfer} and RoboTransfer~\cite{liu2025robotransfer} leverage embodied data to achieve better kinematic plausibility, and we compare \textit{DreamTransfer} with these specialized models.
For fair comparison, we adopt model-specific input-output processing schemes. Cosmos-Transfer1 only supports single-view generation, so we process each camera view separately and concatenate outputs horizontally into multi-view results. By contrast, \textit{DreamTransfer} and RoboTransfer natively handle multi-view inputs, generating complete multi-view videos in one inference step.
All generated videos are at a uniform resolution of $1920 \times 480$ for fair evaluation, matching the platform's hardware (three synchronized $640 \times 480$ cameras) in real-world robot evaluations.

\noindent
\textbf{Evaluation dataset.}
We collected 50 real \texttt{Fold Clothes} demonstrations on the Agilex CobotMagic, and 20 simulation demonstrations for both \texttt{Clean Desk} and \texttt{Throw Bottle} in the NVIDIA Isaac Sim~\cite{NVIDIA_Isaac_Sim}. These tasks were \textbf{unseen} during \textit{DreamTransfer} training, as shown in the first column of Figure~\ref{fig:transfer-vis} (rows 1, 3, and 4).
Scene object appearances stay consistent during data collection.
Each model generates visually diverse demonstrations at a 1:1 ratio with collected data.
All models are evaluated on these generated demonstrations.

\noindent
\textbf{Evaluation metrics.} 
Our evaluation focuses on three key aspects: multi-view consistency, depth consistency, and text-to-video alignment. 
Multi-view consistency is measured via Matched Pixels (Mat.Pix.), calculating average pixel matches between central and left/right wrist-mounted views over all frames. For depth consistency, we evaluate the fidelity of predicted depth against ground truth using Root Mean Squared Error (RMSE), Absolute Relative Error (Abs.Rel.), and Squared Relative Error (Sq.Rel.). CLIP Similarity quantifies alignment between videos and text instructions. 
The specific formulas for these metrics are provided in Appendix~\ref{app:video-generation-quality-evaluation}.

\noindent
\textbf{Results.} 
As shown in Table~\ref{video-transfer-table}, \textit{DreamTransfer} achieves the best average performance across all evaluation metrics.
It improves pixel matching by 42\% over the second-best model, verifying strong cross-view appearance consistency.
For depth consistency, \textit{DreamTransfer} achieves 24\% improvement over Cosmos-Transfer1 and 58\% over RoboTransfer, benefiting from explicit depth conditioning that enforces structural coherence.
Regarding text-video alignment, DreamTransfer achieves the highest average CLIP similarity score, demonstrating competitive semantic alignment.
Visual comparisons are provided in Appendix~\ref{app:video-generation-models-comparison}.
% where we also discuss the limitations of \textit{DreamTransfer}.

\subsection{Real-World Experiments}
\label{sec:real-robot-eval}

We conduct extensive experiments to evaluate our \textit{EMMA} framework on real-world robotic tasks. We focus on three key questions: (1) Can co-training with generated data improve real-world policy performance and generalization to zero-shot object appearances? (2) How does the mixing ratio of real and generated data affect VLA policy performance? (3) Can \textit{AdaMix}, our hard-sample-aware adaptive training strategy, further enhance real-world policy performance?

\noindent
\textbf{Implementation details.} 
We adopt $\pi_0$~\cite{black2024pi0} and $\pi_{0.5}$~\cite{intelligence2025pi05} as our base VLA policy model architecture without modifications and perform post-training on the pre-trained model.
We evaluate our framework on three challenging real-world robotic tasks: \texttt{Fold Clothes}, \texttt{Clean Desk}, and \texttt{Throw Bottle}. Detailed descriptions of each task are provided in Appendix~\ref{app:task-setup}.

We train policies on a mixed dataset $D^{\alpha}$, composed of real data $D_R$ and generated data $D_G$. 
No.Aug. in Table~\ref{tab:combined_performance} denotes training using real data only, where $|D_R| = 50$ for \texttt{Fold Clothes} and $|D_R| = 20$ for both \texttt{Clean Desk} and \texttt{Throw Bottle}.
The next two rows correspond to training with a 1:1 mixture of real data and data generated by Cosmos-Transfer1~\cite{nvidia2025cosmostransfer} and \textit{DreamTransfer}, as described in Section~\ref{sec:video-generation-quality}. $|D_G| = |D_R| = 50$ for \texttt{Fold Clothes}, and $|D_G| = |D_R| = 20$ for both \texttt{Clean Desk} and \texttt{Throw Bottle}.
These training data settings apply to all experiments unless otherwise specified.
For each task, the real data is collected in a single environment with a fixed set of foreground objects. 
For fair comparison, all experiments for a given task share the same training configuration.
More detailed training configurations are provided in Appendix~\ref{app:vla-training-config}.

We use both task success rate and a behavior score to evaluate policy performance. The behavior score assesses execution quality by decomposing tasks into critical stages, with normalized scores in $[0, 5]$.
Detailed rules are in Appendix~\ref{app:evaluation-details}.
Across all tasks, the visual appearances of the foreground objects in the evaluation environments are entirely distinct from those in the collected real data.
Each reported result is averaged over 10 trials and 4 distinct foreground objects with \textbf{zero-shot} appearances, totaling 40 trials per setting.

\noindent
\textbf{Experimental platform.} 
Experiments run on an Agilex CobotMagic platform with two PiPER arms and three Intel RealSense D435i cameras (two wrist-mounted, one head-mounted). 

\subsubsection{Comparison with Baseline Generation Models and Generalization Tests}
\label{sec:compare_baselines}

As shown in Table~\ref{tab:combined_performance}, among the first three rows under each VLA, training without data augmentation yields the lowest performance across all tasks, proving generated data vital for generalization to unseen visuals.
Policies augmented with Cosmos-Transfer1~\cite{nvidia2025cosmostransfer} show moderate improvements yet struggle with deformable object manipulation, reflected by low success rate on \texttt{Fold Clothes}.
In contrast, \textit{DreamTransfer} achieves consistent and substantial improvements across all three tasks, outperforming both the no-augmentation baseline and Cosmos-Transfer1~\cite{nvidia2025cosmostransfer}. Notably, on \texttt{Fold Clothes}, \textit{DreamTransfer} achieves 65\% and 75\% success rates under $\pi_0$ and $\pi_{0.5}$ respectively, nearly doubling the success rate of the non-augmented baseline. This demonstrates that multi-view consistent, geometrically plausible video generation enables more effective policy learning, especially for challenging tasks involving non-rigid dynamics.

\subsubsection{Impact of Generated Data Mixing Ratio on Real-World Robot Performance}
\label{sec:exp_mix_ratio}

\begin{figure*}[h]
    \centering
    \begin{subfigure}[b]{0.32\columnwidth}
        \centering
        \includegraphics[width=\textwidth]{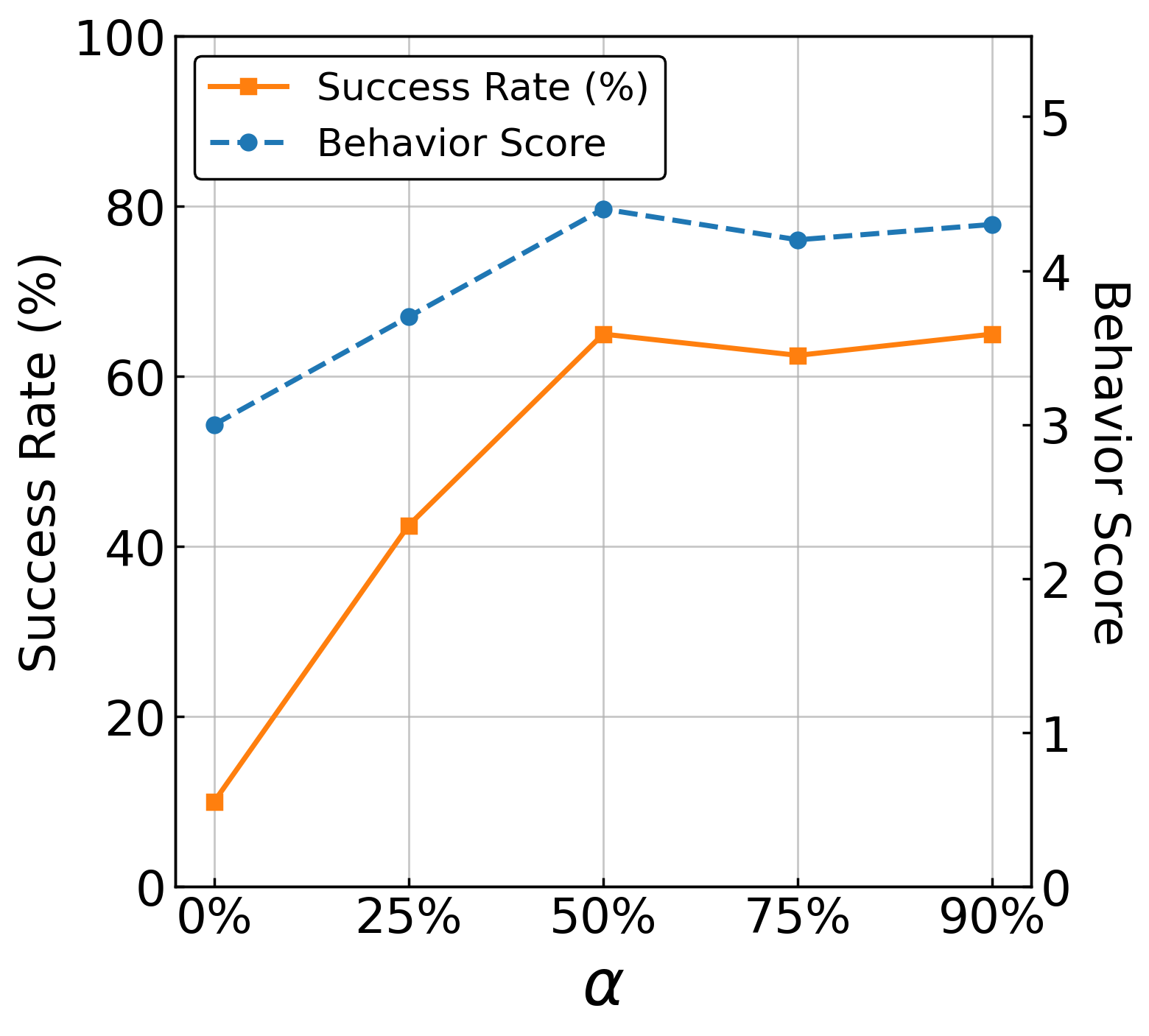}
        \caption{Fold Clothes}
        \label{fig:fold-cloth-mix-ratio}
    \end{subfigure}
    \hfill
    \begin{subfigure}[b]{0.32\columnwidth}
        \centering
        \includegraphics[width=\textwidth]{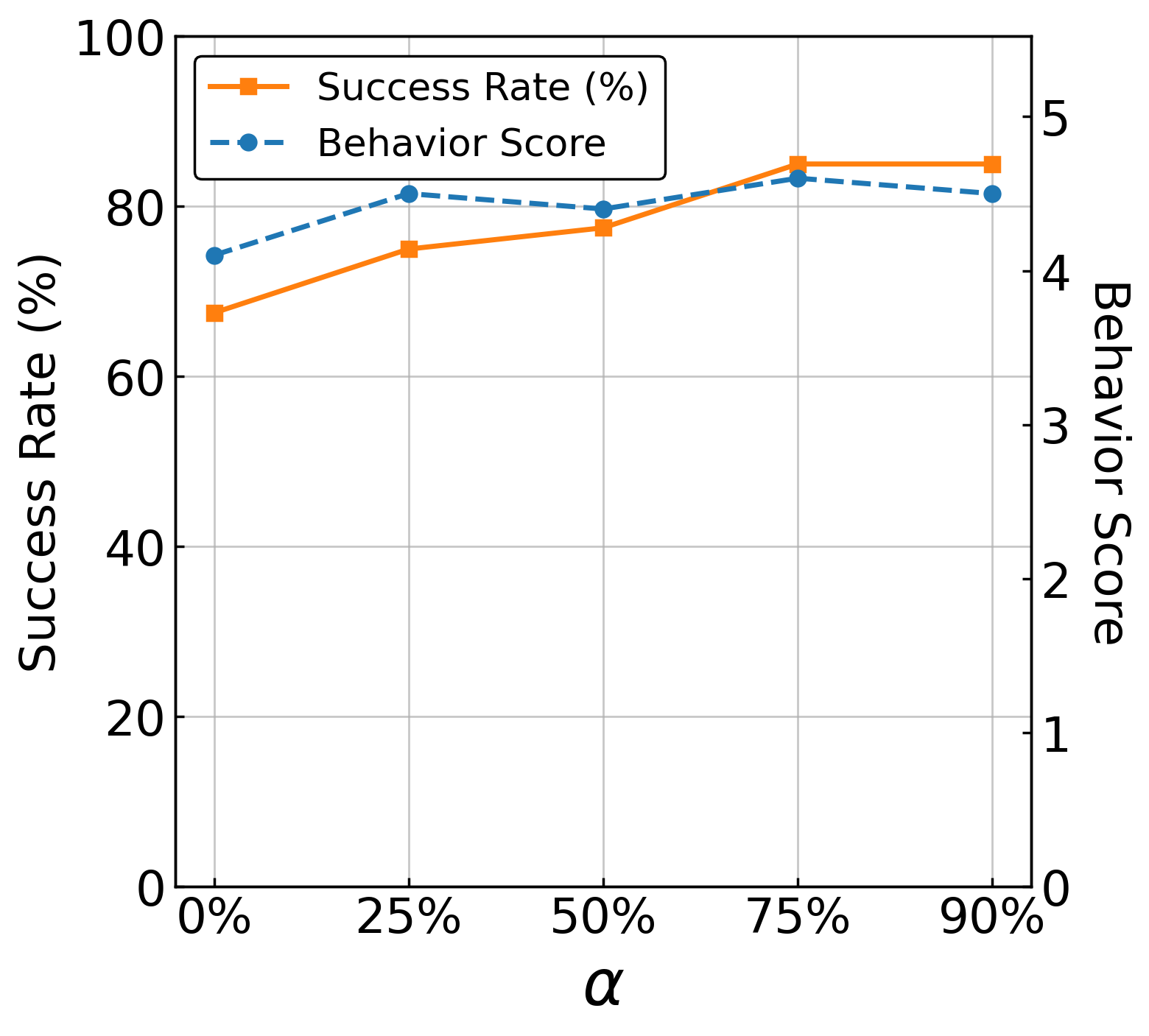}
        \caption{Clean Desk}
        \label{fig:clean-desk-mix-ratio}
    \end{subfigure}
    \hfill
    \begin{subfigure}[b]{0.32\columnwidth}
        \centering
        \includegraphics[width=\textwidth]{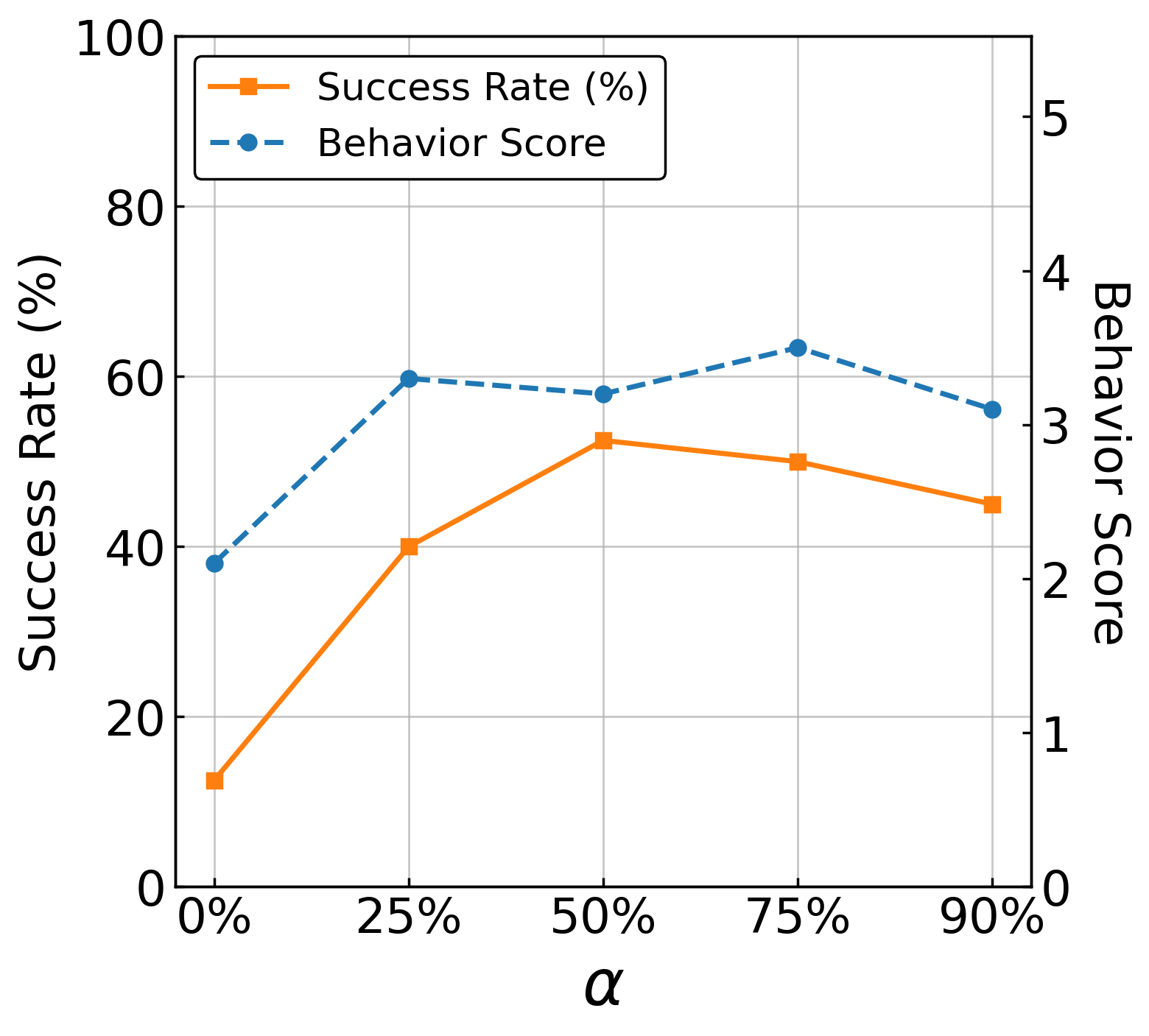}
        \caption{Throw Bottle}
        \label{fig:throw-bottle-mix-ratio}
    \end{subfigure}
    
    \caption{Impact of data mix ratios on real-world robotic tasks performance.}
    \label{fig:data-mix-ratio}
\end{figure*}

We study how the mixing ratio between real and generated data affects policy performance, while keeping the total amount of training data and the number of training steps fixed across all mixing ratios for a given task. 
During training, each training sample is drawn from $D_G$ with probability $\alpha$ and from $D_R$ otherwise, where $\alpha$ serves as the data mixing ratio~\cite{wei2025empirical}. 

As shown in Figure~\ref{fig:data-mix-ratio}, performance improves significantly when generated data is introduced, peaking at a 50\% mixing ratio. This balanced mix achieves optimal generalization, particularly on appearance-sensitive tasks like \texttt{Fold Clothes}, where success rate jumps from 10\% (real data only) to 65\%. Beyond 50\%, performance plateaus: increasing the proportion of generated data to 75\% or 90\% yields no further gain in average success rate and slightly reduces consistency. Notably, at 90\% generated data, the \texttt{Throw Bottle} task exhibits a drop in success rate. This suggests that excessive reliance on generated data may propagate subtle visual or dynamic inaccuracies that harm performance in fast, precision-critical tasks.
This result also highlights the need for the video quality filter in our \textit{EMMA} framework to remove visually or physically unrealistic demonstrations.

\begin{table}[h]
\caption{Comprehensive real-world robotic task performance comparison across video generation models, and training strategies. 
No.Aug. denotes training on real data only. 
$\dagger$ denotes that the generated data in the training set are used without applying the video quality filter.
Both FixMix and \textit{AdaMix} employ a video quality filter before training. 
FixMix uses uniform sampling with constant weights during training.
Score is the behavior score; SR is success rate. Results are averaged over 10 trials across 4 distinct visual variations of the foreground object. 
Best results are in \textbf{bold}.}
\label{tab:combined_performance}
\centering
\resizebox{\linewidth}{!}{%
\begin{tabular}{l @{\hspace{1em}} l c c c c c c c c}
\toprule
\multirow{2}{*}{VLA} & \multirow{2}{*}{Method}
& \multicolumn{2}{c}{Fold Clothes}
& \multicolumn{2}{c}{Clean Desk}
& \multicolumn{2}{c}{Throw Bottle}
& \multicolumn{2}{c}{Average} \\ 
\cmidrule(lr){3-4} \cmidrule(lr){5-6} \cmidrule(lr){7-8} \cmidrule(lr){9-10}
& & Score & SR & Score & SR & Score & SR & Score & SR \\
\midrule

\multirow{5}{*}{\textbf{$\pi_{0}$}~\cite{black2024pi0}}
& No.Aug.   
& 3 & 10\% & 4.1 & 67.5\% & 2.1 & 12.5\% & 3.1 & 30\% \\

& Cosmos-Transfer1$^\dagger$~\cite{nvidia2025cosmostransfer}
& 3.3 & 40\% & 4 & 70\% & 3.2 & 40\% & 3.5 & 50\% \\

& \textit{DreamTransfer}$^\dagger$ 
& 4.4 & 65\% & 4.4 & 77.5\% & 3.2 & 52.5\% & 4 & 65\% \\

& \textit{DreamTransfer}+FixMix
& 4.4 & 70\% & 4.5 & 77.5\% & 3.6 & 57.5\% & 4.2 & 68.3\% \\

& \cellcolor{lightgray}\textit{DreamTransfer}+\textit{AdaMix} 
& \cellcolor{lightgray}\textbf{4.6} & \cellcolor{lightgray}\textbf{77.5\%} & \cellcolor{lightgray}\textbf{4.7} & \cellcolor{lightgray}\textbf{90\%} & \cellcolor{lightgray}\textbf{4.4} & \cellcolor{lightgray}\textbf{70\%} & \cellcolor{lightgray}\textbf{4.6} & \cellcolor{lightgray}\textbf{79.2\%} \\

\midrule

\multirow{5}{*}{\textbf{$\pi_{0.5}$}~\cite{intelligence2025pi05}}
& No.Aug.   
& 3.4 & 27.5\% & 4.4 & 70\% & 2.7 & 32.5\% & 3.5 & 43.3\% \\

& Cosmos-Transfer1$^\dagger$~\cite{nvidia2025cosmostransfer}
& 4.1 & 67.5\% & 4.6 & 77.5\% & 3.1 & 45\% & 3.9 & 63.3\% \\

& \textit{DreamTransfer}$^\dagger$ 
& 4.4 & 75\% & 4.7 & 85\% & 3.7 & 60\% & 4.3 & 73.3\% \\

& \textit{DreamTransfer}+FixMix 
& 4.5 & 77.5\% & 4.6 & 87.5\% & 3.6 & 60\% & 4.2 & 75\% \\

& \cellcolor{lightgray}\textit{DreamTransfer}+\textit{AdaMix} 
& \cellcolor{lightgray}\textbf{4.6} & \cellcolor{lightgray}\textbf{82.5\%} & \cellcolor{lightgray}\textbf{4.8} & \cellcolor{lightgray}\textbf{92.5\%} & \cellcolor{lightgray}\textbf{4.2} & \cellcolor{lightgray}\textbf{72.5\%} & \cellcolor{lightgray}\textbf{4.5} & \cellcolor{lightgray}\textbf{82.5\%} \\

\bottomrule
\end{tabular}
}
\end{table}

\subsubsection{Ablation Study on AdaMix Training Framework}
We conduct ablation experiments to verify \textit{AdaMix}'s effectiveness in improving real-world policy performance. 
All models are trained with identical steps and mixed real-generated datasets for each task.
Training starts with unified initialization: all samples that pass the video quality filter are assigned equal initial weights, while low-quality samples are assigned zero weight and excluded from training. 
Low-quality data accounts for 8\% in \texttt{Fold Clothes}, 5\% in both \texttt{Clean Desk} and \texttt{Throw Bottle}.
The key difference lies in whether sampling weights are adapted during training.

\begin{table*}[h]
\caption{Comparison of execution time (Time, seconds), trajectory smoothness (Smth., angular acceleration in rad$/s^2$) and joint overlimit (JOL., frames) between FixMix and \textit{AdaMix} training strategies on real-world robotic tasks. Detailed calculations for these metrics are provided in Appendix~\ref{app:evaluation-details}. Lower values are better for all metrics. Best results are in \textbf{bold}.}
\label{tab:execution-performance}
\centering
\resizebox{0.9\linewidth}{!}{
\begin{tabular}{l c c c c c c c c c c c c}
\toprule
\multirow{2}{*}{Method} & \multicolumn{3}{c}{Fold Clothes} & \multicolumn{3}{c}{Clean Desk} & \multicolumn{3}{c}{Throw Bottle} & \multicolumn{3}{c}{Average} \\ 
\cmidrule(lr){2-4} \cmidrule(lr){5-7} \cmidrule(lr){8-10} \cmidrule(lr){11-13}
& Time & Smth. & JOL. & Time & Smth. & JOL. & Time & Smth. & JOL. & Time & Smth. & JOL. \\
\midrule
FixMix        & 40.4 & \textbf{2.2} & 56.1 & 12.1 & 2.6 & \textbf{60.0} & 46.0 & \textbf{1.2} & 14.6 & 32.8 & 2.0 & 43.6 \\
\textit{AdaMix} & \textbf{39.8} & \textbf{2.2} & \textbf{53.4} & \textbf{11.3} & \textbf{2.4} & 64.9 & \textbf{38.9} & \textbf{1.2} & \textbf{10.4} & \textbf{30.0} & \textbf{1.9} & \textbf{42.9} \\
\bottomrule
\end{tabular}%
}
\end{table*}

In the FixMix baseline, sampling weights stay fixed for uniform sampling of retained data.
In contrast, \textit{AdaMix} dynamically updates the sampling weights after the first half of training steps, prioritizing challenging and informative samples via trajectory performance scores (see Section~\ref{sec:adamix} for details).
\textit{AdaMix} is applied jointly to both real and generated data for hard sample identification, and its key strength is being seamlessly compatible with generated and real data during training. 

As summarized in Table~\ref{tab:combined_performance}, \textit{AdaMix} consistently improves task performance across all VLA policies.
Compared to FixMix, it increases the average success rate by 10.9 and 7.5 percentage points for $\pi_0$ and $\pi_{0.5}$, respectively.
\textit{AdaMix} also improves overall execution quality compared with FixMix, as shown in Table~\ref{tab:execution-performance}.
VLA policies trained with FixMix often struggle with imprecise grasps and repeated adjustments. In contrast, \textit{AdaMix} achieves better grasping precision and motion stability to avoid such flaws.

\subsection{Simulation Experiments}
\label{sec:simulation-experiments}

\begin{wrapfigure}{r}{5cm}
    \centering 
    \includegraphics[width=\linewidth]{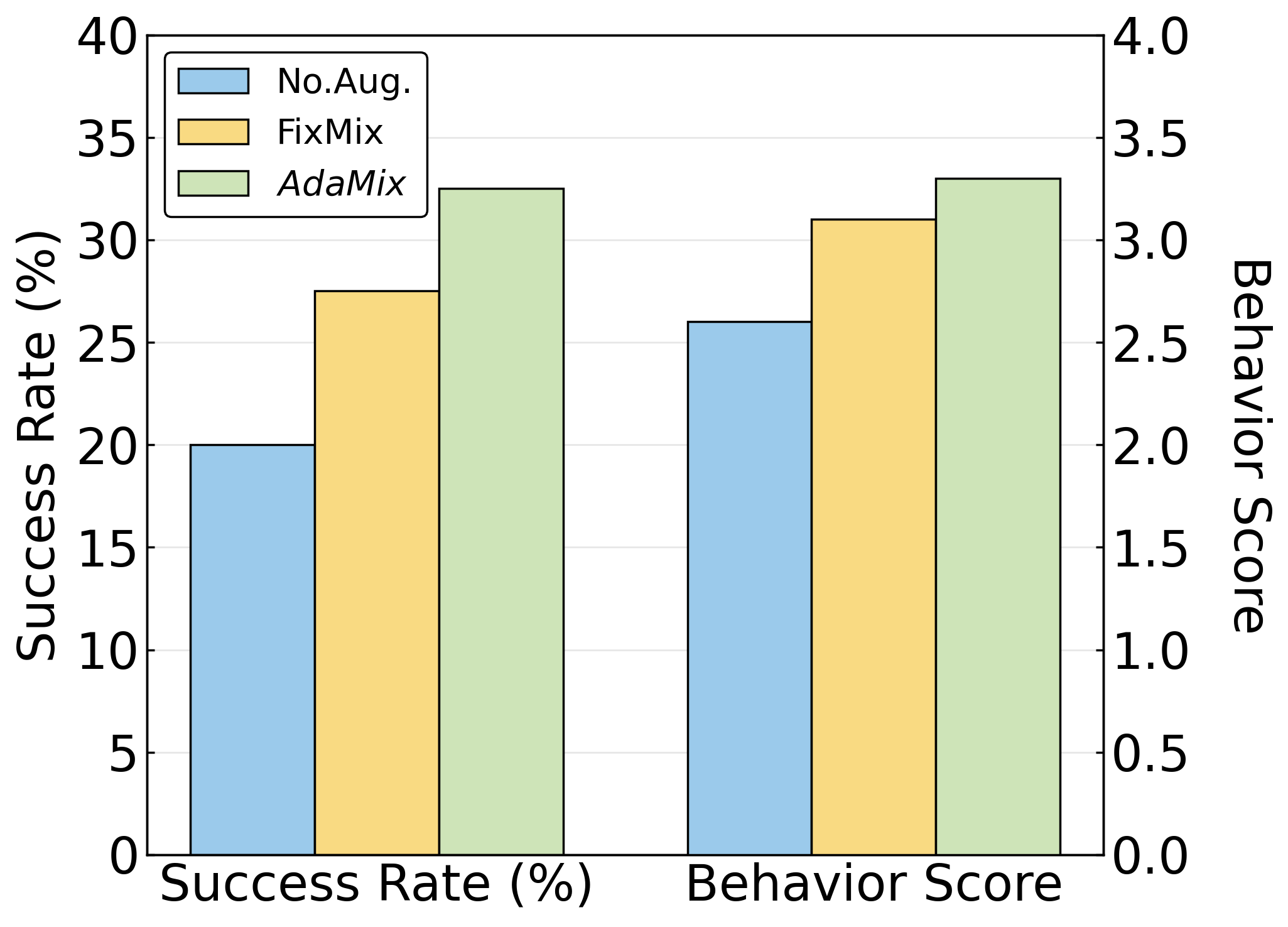}
    \caption{Task success rate and behavior score under different training data and mixing strategies in simulation experiments.
    }
    \label{fig:sim-score-success-rate}
\end{wrapfigure}

\textbf{Simulation setup.}
We conduct simulation experiments in the RoboTwin~2.0~\cite{chen2025robotwin20scalabledata} on the \texttt{Stack Bowls Two} task. We collect 50 original demonstration trajectories, where the visual appearance of objects is consistent across different demonstrations. 
\textit{DreamTransfer} generates another 50 demonstrations with diverse object appearances at a 1:1 data ratio.

We consider three training settings: (i)~No.Aug., training on the original simulation demonstrations; (ii)~FixMix, training on both original and generated demonstrations with uniform sampling weights after applying the video quality filter (6\% of the data is filtered out); and (iii)~\textit{AdaMix}, training on the same filtered dataset but updates sampling weights adaptively during training based on trajectory performance scores. All policies are trained using the $\pi_0$ architecture and training configurations detailed in Appendix~\ref{app:vla-training-config}.
For evaluation, each policy runs 40 trials in environments with bowls of unseen visual appearances. 
The behavior score rules are provided in Appendix~\ref{app:evaluation-details}.

\noindent
\textbf{Results.}
As shown in Fig~\ref{fig:sim-score-success-rate} and Table~\ref{tab:sim-performance}, training with additional data generated by \textit{DreamTransfer} consistently improves task success rate compared with using only original demonstrations. Moreover, the proposed \textit{AdaMix} strategy further enhances performance over the baseline mixing scheme. These observations align with our real-world robot evaluations.

%% file: sec/5-conclusion.tex
\section{Conclusion}
\label{conclusion}

In this paper, we address the challenge of scaling up diverse and generalizable VLA learning for robot manipulation, where real-world data collection is costly and simulation lacks realism. 
We present \textit{EMMA}, a framework for enhancing VLA policy via text-guided embodied manipulation video generation with \textit{DreamTransfer} and adaptive training with \textit{AdaMix}.
Evaluated on real-world tasks under zero-shot visual conditions, \textit{EMMA} achieves over 92\% improvement in success rate compared to training on real data alone, with an additional 17\% gain from \textit{AdaMix}. 
These results demonstrate that \textit{EMMA} provides an effective way to enhance the generalization of VLA.

\section{Limitations}
Although our method achieves high success rate, it is limited when handling tiny objects in cluttered scenarios. Video depth estimation models~\cite{chen2025videodepthanything} produce inaccurate depth maps and artifacts on tiny objects, which impairs the training of the downstream VLA policy. This highlights the need for more advanced depth estimation models to improve robotic manipulation in complex real-world scenes.

%% file: sec/6-appendix.tex
\appendix

\section{Related Work}
\label{app:related-work}

\subsection{Visual Generalizable Imitation Learning}

Imitation learning enables visuomotor policies to learn from human demonstrations, providing an effective pathway for robotic manipulation~\cite{chi2024diffusionpolicy,li2025roboticmanipulationimitationlearning,jiang2025rynnvla001}. 
Recently, VLA models have significantly improved generalization by integrating semantic understanding into action generation~\cite{brohan2023rt2,kim2024openvla}. 
Early systems like CLIPort~\cite{shridhar2021cliport} established the foundation for vision-conditioned control, while subsequent works enhanced capabilities through chain-of-thought reasoning ~\cite{zhen2025tesseract}. Recent advances include domain specialization~\cite{yue2024deervla}, occlusion handling~\cite{wei2024occllama}, and safety-aware execution~\cite{zhang2025safevla}. Models like OpenVLA and EF-VLA~\cite{huang2025early} further improve performance through dual visual encoders or preserved semantic alignment, while self-correcting frameworks~\cite{li2025selfcorrecting} enable recovery from failures in cluttered environments.
However, VLA models perform poorly under limited real-world demonstrations and require large-scale and diverse data to generalize across objects and environments.

Collecting data across diverse objects and environments is expensive and time-intensive~\cite{embodimentcollaboration2025openx,khazatsky2025droid}. This gap motivates using generated data as a scalable way to enrich visual diversity, provided that the generated content maintains geometric plausibility and spatial consistency.

\subsection{Generative Models for Embodied Data}

To improve VLA model generalization, various generative approaches have been proposed for low-cost diverse robot data generation~\cite{lin2025data,chen2024roviaugrobot,jin2025physically,yuan2025motiontranshumanvrdata,wang2025embodiedreamer}.
Traditional data augmentation works for in-domain generalization~\cite{chi2024diffusionpolicy} but struggles under large visual distribution shifts. 
Generative models offer stronger cross-domain adaptation~\cite{teoh2024greenscreen}, yet many rely on additional inputs like object masks or scene annotations~\cite{chen2024semantically,mandi2023cacti,wang2024cyberdemo}.
Inpainting-based techniques often require per-scene tuning~\cite{zhuang2024enhancing,yu2023scalingrobotlearning}.
Real-to-sim-to-real methods like RoboSplat~\cite{yang2025noveldemonstrationgeneration} and ReBot~\cite{fang2025rebotscalingrobotlearning} rely on simulators and are largely limited to rigid-body interactions.
Recent advances in diffusion models~\cite{ho2020denoising,song2022denoising,ho2022classifierfree,blattmann2023stablevideodiffusion} and video diffusion transformers~\cite{lu2023vdtgeneral,yang2025cogvideox} have enabled high-fidelity generation of embodied manipulation videos.
However, they lack text-driven control over scene appearance while preserving cross-view consistency. 
Cosmos-Transfer1~\cite{nvidia2025cosmostransfer} uses semantic segmentation and depth maps to generate realistic scenes. RoboTransfer~\cite{liu2025robotransfer} improves multi-view consistency using depth and surface normals, but relies on image inputs and does not support text-based editing of manipulation videos.
Existing multi-view diffusion models (MVDream~\cite{shi2024mvdream}, SPAD~\cite{kant2024spad}, Zero-1-to-3~\cite{liu2023zero}) target static 3D modeling and lack temporal modeling for dynamic robot manipulation. In contrast, \textit{DreamTransfer} is tailored for manipulation videos, combining depth and temporal modeling for interactive dynamics. While prior work stops at visual generation, our framework leverages \textit{AdaMix} to enable end-to-end enhancement of VLA policy robustness.

\begin{figure*}[!htbp]
    \centering
    \includegraphics[width=\textwidth]{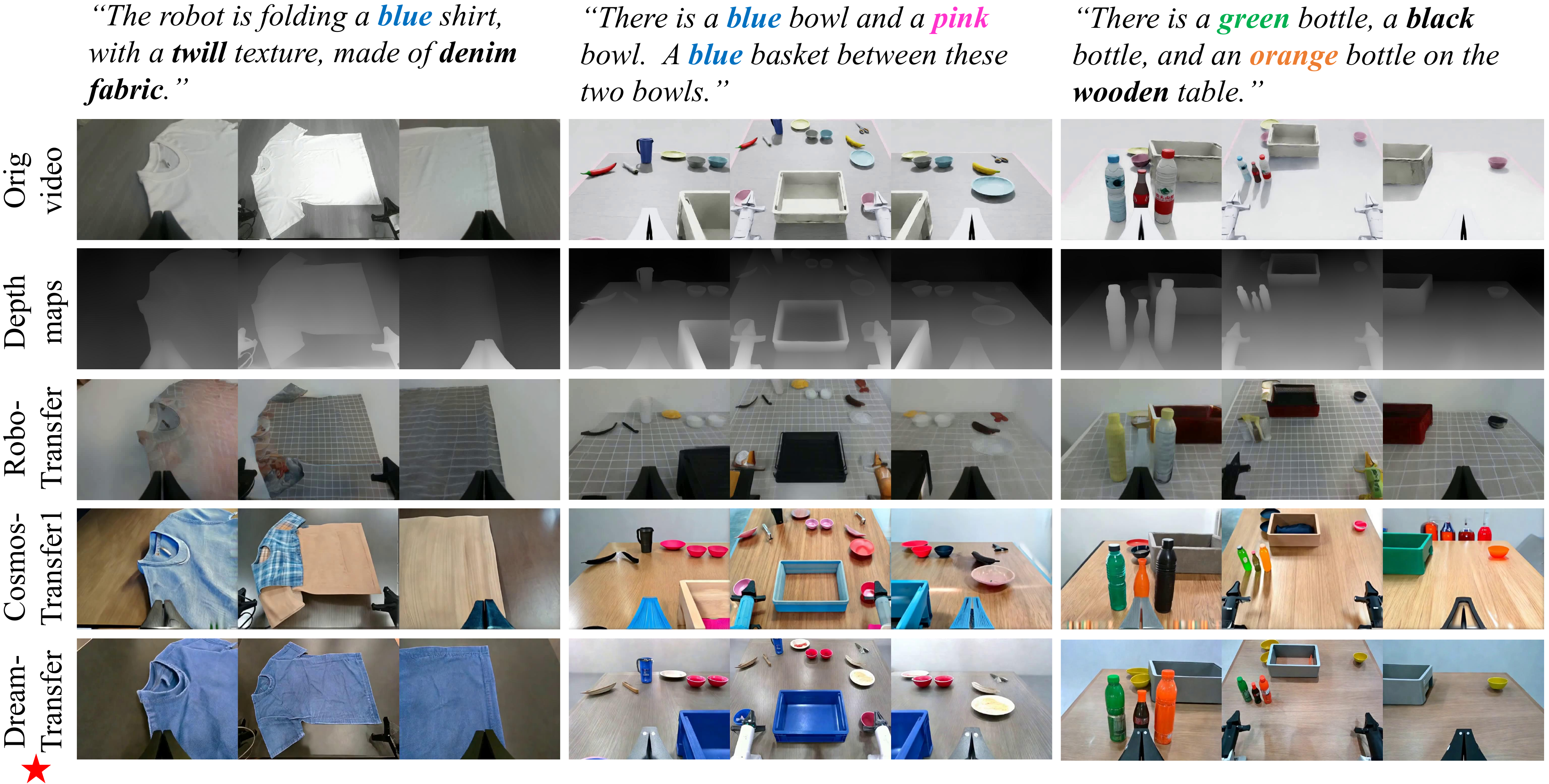}
    \caption{Qualitative comparison of embodied manipulation video transfer methods.}
    \label{fig:transfer-compare-demo}
\end{figure*}

\section{More Experiments}
\label{app:more_experiments}

In this section, we present additional experimental results to further validate our proposed method. Unless otherwise specified, the training configurations for the \texttt{Clean Desk} and \texttt{Throw Bottle} tasks strictly follow the settings detailed in their respective columns of Table~\ref{tab:vla-training-config}. Furthermore, the $\pi_0$ policy is adopted as the base architecture for all evaluations presented here, including the hyperparameter analysis and the ablation study on trajectory quality metrics (which supplements the Real-World Robot Evaluation section in the main paper).

\subsection{Ablation Study on the Trajectory Quality Metrics}
\label{app:ablation_metrics}

We conduct an ablation study to evaluate the individual contribution of each metric used in computing the sample difficulty in Equation~6 of the main paper. 
The evaluation is performed on the \texttt{Throw Bottle} task.
The results, summarized in Table~\ref{tab:ablation_metrics_throw}, show that removing any one metric leads to a consistent decline in both success rate and behavior score, and no single metric dominates performance across all conditions. Specifically, action MSE has the strongest individual impact, but smoothness and joint angle limitation still provide complementary benefits that improve overall robustness. 

\begin{table}[h]
\centering
\begin{minipage}{0.53\linewidth}
\centering
\caption{Ablation study on trajectory quality metrics. Each row shows a different metric combination for computing $s_i$.}
\resizebox{\linewidth}{!}{%
\begin{tabular}{ccc|ccc}
\toprule
Action MSE & Smoothness & Joint Limitation & Score & SR \\
\midrule
\checkmark & \checkmark & \checkmark & 4.4 & 70\% \\
\checkmark & \checkmark & & 4.2 & 62.5\% \\
\checkmark & & \checkmark & 3.9 & 55\% \\
& \checkmark & \checkmark & 3.7 & 55\% \\
\bottomrule
\end{tabular}
}
\label{tab:ablation_metrics_throw}
\end{minipage}
\hfill % 两个表格之间添加水平间距，确保排版整齐
% 第二个表格：占据页面宽度的45%
\begin{minipage}{0.45\linewidth}
\centering
\caption{Trajectory quality metrics of the \texttt{Stack Bowls Two} task (Simulation Experiments in main paper). Comparison between FixMix and \textit{AdaMix} training strategies. Best results are in \textbf{bold}.}
\label{tab:sim-performance}
\resizebox{\linewidth}{!}{%
\begin{tabular}{l c c c}
\toprule
Method & Time & Smoothness & Joint Overlimit \\ 
\midrule
FixMix & 12.7 & \textbf{1.9} & 59.1 \\
\textit{AdaMix} & \textbf{12.3} & \textbf{1.9} & \textbf{43.4} \\
\bottomrule
\end{tabular}
}
\end{minipage}
\end{table}

\subsection{\texorpdfstring{Hyperparameter Analysis of $\gamma$ and $\lambda$}{Hyperparameter analysis of gamma and lambda}}
\label{app:study_hyperpara}

% \begin{wrapfigure}{r}{5cm}
%     \vspace{-2em}
%     \centering 
%     \includegraphics[width=\linewidth]{figures/study_lambda.png}
%     \caption{Performance of the policy under different $\lambda$ with fixed $\gamma = 1$.}
%     \label{fig:study-lambda}
% \end{wrapfigure}

We analyze the impact of the hyperparameters $\gamma$ and $\lambda$ in the sampling weight defined in Equation~7 of the main paper. The evaluation is conducted on the \texttt{Throw Bottle} task.
We fix $\gamma = 1$ to ensure all samples receive non-zero sampling probability and vary $\lambda \in \{1, 10, 20, 100\}$. 
The result is shown in Figure~\ref{fig:study-lambda}.
When $\lambda = 1$, the policy performs similarly to uniform sampling, indicating insufficient focus on hard examples. Larger values of $\lambda$ degrade performance. This suggests that over-prioritizing hard samples reduces exposure to informative easy cases and amplifies noise or outliers.
These results indicate that $\lambda = 10$ provides an effective trade-off between hard-sample focus and data diversity, while $\gamma = 1$ ensures baseline coverage. 

\subsection{Impact of Real Data Volume vs. Data Quality}
\label{sec:data_volume}

% \begin{wrapfigure}{r}{5cm}
%     \centering 
%     \includegraphics[width=\linewidth]{figures/real_data_size_effect.png}
%     \caption{Evaluation of the effect of real data volume on the \texttt{Clean Desk} task.}
%     \label{fig:real-data-size-effect}
% \end{wrapfigure}

\begin{figure*}[htbp]
    \centering
    \begin{subfigure}[b]{0.45\columnwidth}
        \centering
        \includegraphics[width=\textwidth]{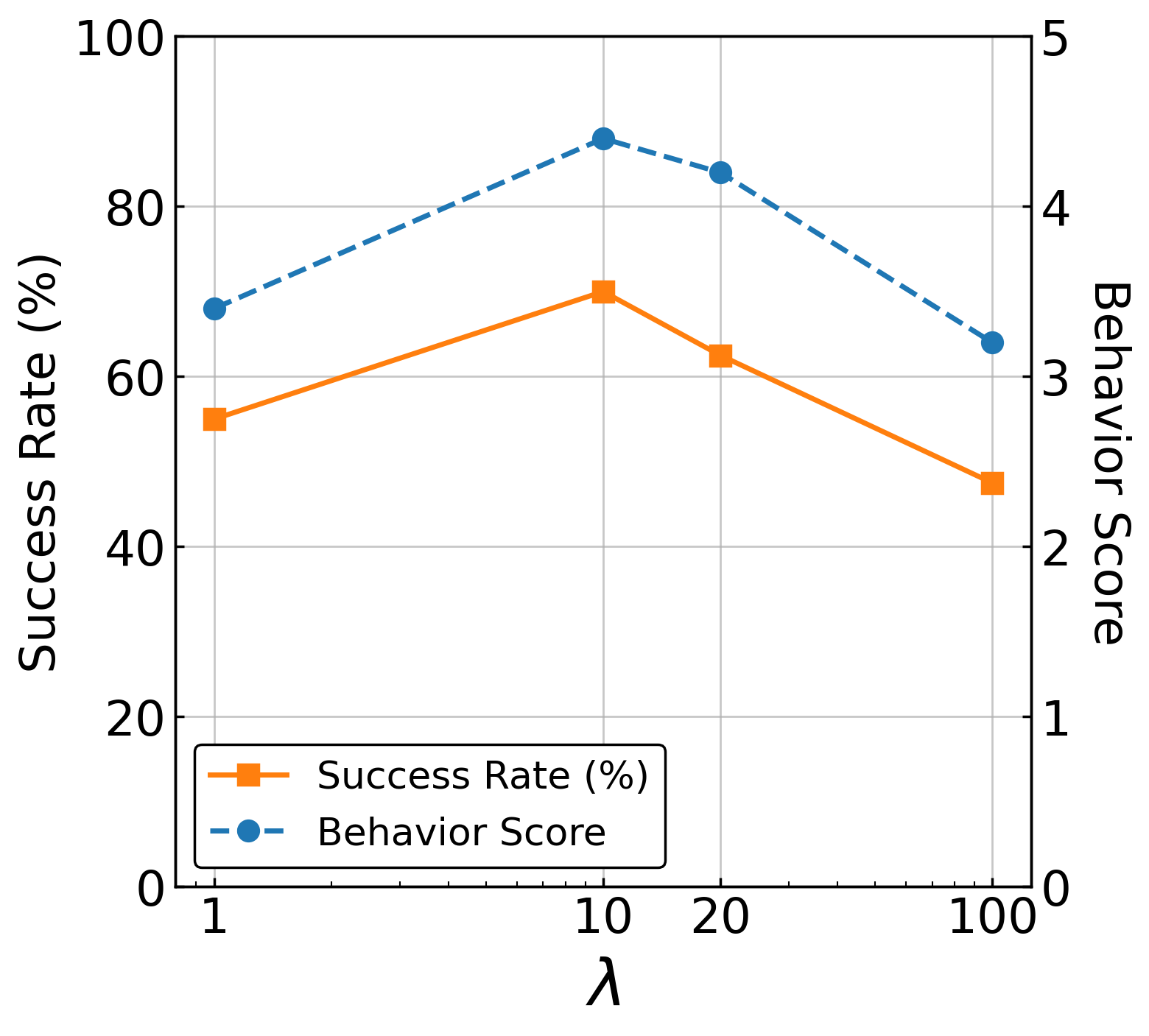}
        \caption{Performance of the policy under different $\lambda$ with fixed $\gamma = 1$.}
        \label{fig:study-lambda}
    \end{subfigure}
    \hfill
    \begin{subfigure}[b]{0.45\columnwidth}
        \centering
        \includegraphics[width=\textwidth]{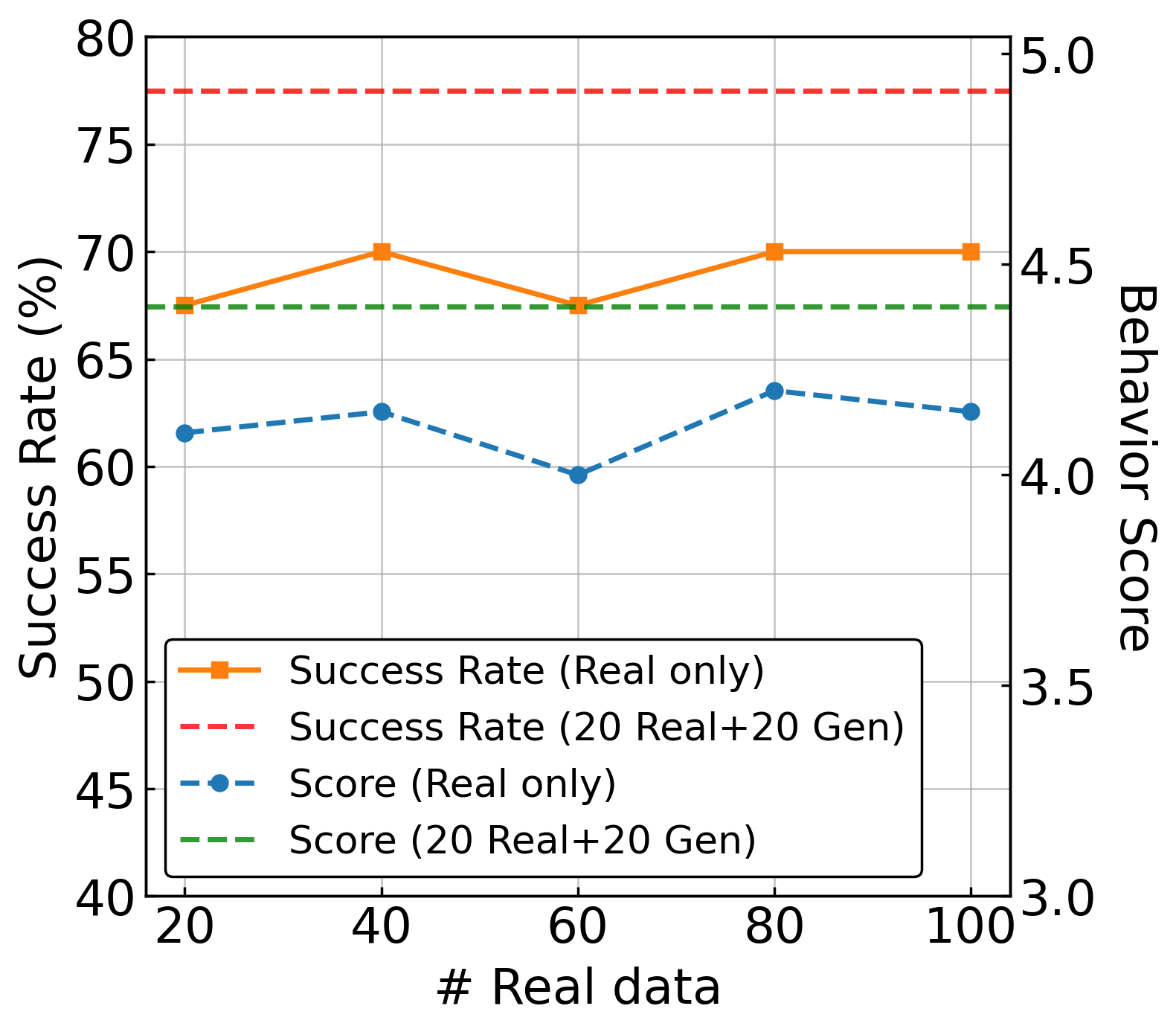}
        \caption{Evaluation of the effect of real data volume on the \texttt{Clean Desk} task.}
        \label{fig:real-data-size-effect}
    \end{subfigure}
    \caption{Ablation studies on the sampling hyperparameter $\lambda$ and the effect of real data volume.}
    \label{fig:two_figs_together}
\end{figure*}

We investigate whether the performance improvements brought by \textit{DreamTransfer} originate merely from a larger amount of data or from the generation of more useful and informative data. To this end, we conduct experiments on the \texttt{Clean Desk} task by additionally collecting 80 real-world demonstrations under the same scene and object appearances. We train the $\pi_0$ policy using 20, 40, 60, 80, and 100 real demonstrations, respectively. As illustrated in Figure~\ref{fig:real-data-size-effect}, the results validate that the performance gains primarily arise from more useful and informative data, rather than a simple increase in data volume.

\section{Failure Case Analysis}
\textit{DreamTransfer} generates high-quality videos with strong geometric and multi-view consistency in most scenarios.
However, its depth maps are estimated from RGB videos via Video Depth Anything~\cite{chen2025videodepthanything}, which may fail on fast-moving objects, uniform-color surfaces, reflective materials, transparent objects, or tiny objects. 
Inaccurate depth can cause artifacts, but our video quality filter effectively excludes such failed cases to prevent VLA policy degradation.

\section{Training Details of DreamTransfer}
\label{app:dreamtransfer-train-detail}

\subsection{Dataset Construction}
\label{app:dataset-construction}
\begin{figure*}[!htbp]
    \centering
    \includegraphics[width=\textwidth]{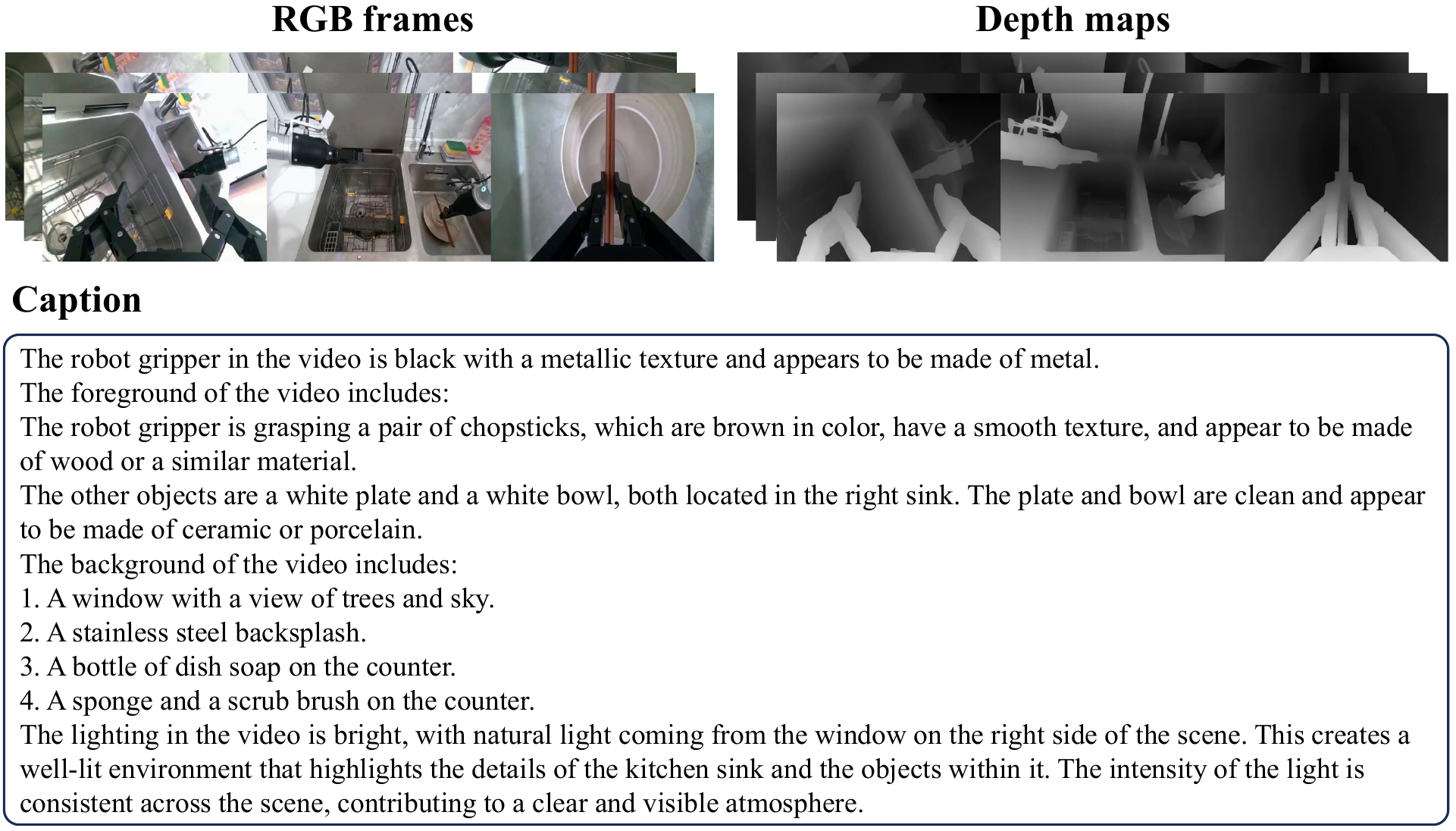}
    \caption{Example from our multi-view embodied manipulation dataset built upon Agibot World dataset. Each sample consists of 121 aligned RGB and depth frames, along with a caption describing object and scene appearance.}
    \label{fig:agibot-sample}
\end{figure*}

To enable multi-view modeling, we construct a dataset of 50k generated multi-view video clips based on the Agibot World dataset~\cite{bu2025agibot}, covering 36 diverse scenes.
This dataset is collected via the AgiBot G1 platform equipped with a standard gripper. By contrast, demonstrations used in our video generation quality experiments are collected from the Agilex CobotMagic platform with two PiPER arms. This further verifies that \textit{DreamTransfer} generalizes to unseen datasets across different robotic arms, environments and tasks.
Each clip has a length of 121 frames and contains aligned multi-view RGB frames, temporally consistent depth maps, and a text caption. Depth maps are generated using the state-of-the-art estimator Video Depth Anything~\cite{chen2025videodepthanything}, ensuring geometric coherence across time. To support fine-grained appearance control, we utilize the prompt template shown below to structure the video captions. This template, which explicitly describes the foreground, background, and lighting conditions, is automatically populated by Qwen2.5-VL-7B-Instruct~\cite{bai2025qwen25vltechnicalreport} to generate detailed, appearance-focused captions.
One sample from the training set is illustrated in Figure~\ref{fig:agibot-sample}.

\begin{tcolorbox}[
colback=blue!5!white,
colframe=blue!5!black,
width=\linewidth,
breakable=true,
title=Prompt template for video caption]
\# Task:\\
Please describe the robot grippers' color, texture and material.\\
Do not return anything other than the appearance of the robot grippers.\\

\# Example: \\
The robot gripper in the video is ... \\

The foreground of the video includes: \\
 - The robot gripper is grasping ... \\
 - The other objects are ... \\

The background of the video includes ... \\
The lighting condition in the video is ...
\end{tcolorbox}

\begin{table*}[h]
\centering
\caption{Training configurations for the two-stage training of \textit{DreamTransfer}.}
\label{tab:training-config}
\begin{tabular}{lcc}
\toprule
\textbf{Configurations} & \textbf{Stage 1} & \textbf{Stage 2} \\
\midrule
Input Resolution (W $\times$ H) & $576 \times 128$ & $1920 \times 480$ \\
Batch Size & 32 & 4 \\
Training Steps & 3500 & 4500 \\
Optimizer & AdamW & AdamW \\
Learning Rate & $1 \times 10^{-5}$ & $1 \times 10^{-5}$ \\
Trainable Parameters & Main Branch + ControlNet & Main Branch + ControlNet \\
Goal & Multi-view consistency & High-res adaptation \\
\bottomrule
\end{tabular}
\end{table*}

\subsection{Training Recipe}
\label{app:training-recipe-dreamtransfer}

We designed \textit{DreamTransfer} to perform appearance transfer directly on multi-view demonstrations from both real-world and simulation environments. Our real-world demonstrations are collected on the AgileX CobotMagic platform, which is equipped with three Intel RealSense D435i cameras, while the simulation demonstrations are collected in the NVIDIA Isaac Sim environment~\cite{NVIDIA_Isaac_Sim}. In both data sources, three synchronized cameras capture RGB images at $640 \times 480$ resolution.
We concatenate the three views along the width to form a $1920 \times 480$ input frame for multi-view consistent modeling.
Accordingly, \textit{DreamTransfer} is trained to generate multi-view videos at $1920 \times 480$ resolution, enabling direct multi-view video generation in one pass.

We fine-tune the pretrained Cosmos-Transfer1 model~\cite{nvidia2025cosmostransfer} on the multi-view embodied manipulation dataset detailed in Appendix~\ref{app:dataset-construction}.
However, the target resolution of $1920 \times 480$ resolution exceeds the maximum resolution supported by the pretrained Cosmos-Transfer1 model, leading to visible artifacts and noise in the generated video. To address this issue with limited computational resources, we adopt a two-stage training strategy. 
\textbf{Stage 1}: We stabilize the learning of multi-view geometric and appearance consistency at $576 \times 128$ resolution.
\textbf{Stage 2}: We progressively adapt the model to $1920 \times 480$ resolution inputs while recovering high-fidelity details in out-of-distribution views.
Through this two-stage training strategy, our model achieves state-of-the-art performance with only 131 hours of training on 4 NVIDIA H20 GPUs. The detailed training configurations for each stage are summarized in Table~\ref{tab:training-config}.

\subsection{Computational Cost and Training Time Comparison}
\label{app:computational-cost}

To ensure a fair comparison with baseline methods, we report the computational resources and training time required by each model in Table~\ref{tab:training-time}. All models are trained to convergence based on their respective protocols.

\begin{table}[h]
\centering
\caption{Training time and computational resources for each model.}
\label{tab:training-time}
\resizebox{0.8\linewidth}{!}{%
\begin{tabular}{lcc}
\toprule
Model & Training Time & GPUs \\
\midrule
Cosmos-Transfer1~\cite{nvidia2025cosmostransfer} & 336 hours & 1024 $\times$ NVIDIA H100 GPUs \\
RoboTransfer~\cite{liu2025robotransfer} & 96 hours & 8 $\times$ NVIDIA H20 GPUs \\
\textit{DreamTransfer} & 131 hours & 4 $\times$ NVIDIA H20 GPUs \\
\bottomrule
\end{tabular}
}
\end{table}

Our two-stage training strategy enables high-quality multi-view video generation with just 131 hours of training on 4 NVIDIA H20 GPUs, requiring significantly fewer GPU hours than both RoboTransfer~\cite{liu2025robotransfer} and Cosmos-Transfer1~\cite{nvidia2025cosmostransfer}.

\section{Video Generation Quality Evaluation and Filtering Metrics}
\label{app:video-generation-quality-evaluation}

\subsection{Video Generation Quality Evaluation}
\label{app:video-quality}

To ensure the realism and fidelity of generated videos for downstream VLA policy training, we employ a set of quantitative evaluation metrics and apply corresponding filtering criteria to filter out low-quality videos. Specifically, we evaluate generated videos using five metrics: matching pixels (Mat.Pix.), root mean squared error (RMSE), absolute relative error (Abs.Rel.), squared relative error (Sq.Rel.), and CLIP similarity (CLIPSim.).

\noindent \textbf{Multi-view consistency.}  
Consistency across camera views is quantified by the number of matching pixels between the center view and the left/right wrist-mounted views. Using a state-of-the-art image matcher~\cite{shen2024gim}, we compute the number of matched pixel correspondences between the center view and each side view for each frame $t$, denoted as $N_{\text{match},t}^{(\text{center},\text{left})}$ and $N_{\text{match},t}^{(\text{center},\text{right})}$. The per-frame matching score is averaged over the two side views, and then averaged over all $T$ frames to obtain the final Mat.Pix. score:
\begin{equation}
    \text{Mat.Pix.} = \frac{1}{T} \sum_{t=1}^T \left[ \frac{1}{2} \left( N_{\text{match},t}^{(\text{center},\text{left})} + N_{\text{match},t}^{(\text{center},\text{right})} \right) \right].
\end{equation}

\noindent \textbf{Depth consistency.}  
To evaluate 3D structural plausibility, we extract temporally coherent depth maps from both original and generated videos using Video Depth Anything~\cite{chen2025videodepthanything}. Let $\mathbf{D}_v \in \mathbb{R}^{T \times H \times W}$ and $\hat{\mathbf{D}}_v \in \mathbb{R}^{T \times H \times W}$ denote the ground-truth and predicted depth tensors for view $v \in \{\text{center}, \text{left}, \text{right}\}$, respectively. We flatten these tensors into vectors $\mathbf{d}_v, \hat{\mathbf{d}}_v \in \mathbb{R}^{N}$ with $N = T H W$, and compute the following scale-invariant error metrics:

\begin{equation}
    \text{RMSE}_v = \frac{1}{\sqrt{N}} \left\| \hat{\mathbf{d}}_v - \mathbf{d}_v \right\|_2,
\end{equation}
\begin{equation}
    \text{Abs.Rel.}_v = \frac{1}{N} \sum_{n=1}^{N} \left| \frac{ \hat{d}_{v,n} - d_{v,n} }{ d_{v,n} } \right|,
\end{equation}
\begin{equation}
    \text{Sq.Rel.}_v = \frac{1}{N} \sum_{n=1}^{N} \left( \frac{ \hat{d}_{v,n} - d_{v,n} }{ d_{v,n} } \right)^2,
\end{equation}
where $d_{v,n}$ and $\hat{d}_{v,n}$ are the $n$-th elements of $\mathbf{d}_v$ and $\hat{\mathbf{d}}_v$, respectively. The final metric values are obtained by averaging across the three views:
\begin{equation}
    \text{Metric} = \frac{1}{3} \left( \text{Metric}_{\text{center}} + \text{Metric}_{\text{left}} + \text{Metric}_{\text{right}} \right),
\end{equation}
for $\text{Metric} \in \{\text{RMSE}, \text{Abs.Rel.}, \text{Sq.Rel.}\}$.

\noindent \textbf{Text-video alignment.}  
Semantic fidelity to the input prompt is measured via CLIP similarity~\cite{radford2021clip}. Given a text prompt that specifies the foreground object, background scene, and lighting condition, we encode the prompt into a text embedding $\mathbf{t}$ and compute the CLIP image embedding $\mathbf{i}_t$ for each frame $t$ of the center-view video. The overall CLIP similarity score is computed as the temporal average of per-frame cosine similarities scaled by $\exp(s)$, where $s$ is the learned logit-scale parameter of CLIP:
\begin{equation}
    \text{CLIPSim.} = \frac{\exp(s)}{T} \sum_{t=1}^{T} \frac{ \mathbf{i}_t^\top \mathbf{t} }{ \|\mathbf{i}_t\| \|\mathbf{t}\| }.
\end{equation}

\subsection{Details of the Video Quality Filter in AdaMix}
To ensure high-quality training data for downstream vision-language-action (VLA) policy learning, we apply a strict multi-criteria filtering mechanism based on the five video quality metrics introduced in Section~\ref{app:video-quality}. For each generated video, we evaluate its Mat.Pix., RMSE, Abs.Rel., Sq.Rel., and CLIP similarity scores. Only videos that simultaneously satisfy all threshold conditions specified in Table~\ref{tab:video_quality_thresholds} are retained for training. Videos failing any single criterion are effectively excluded by setting their sampling probability to zero.

\begin{table}[h]
\centering
\caption{Quality thresholds for filtering generated videos in \textit{AdaMix}. A video is retained only if all its metric values meet the corresponding criteria.}
\label{tab:video_quality_thresholds}
\resizebox{0.6\linewidth}{!}{%
\begin{tabular}{l|ccccc}
\toprule
Metric & Mat.Pix. & RMSE & Abs.Rel. & Sq.Rel. & CLIPSim. \\
\midrule
Threshold & $> 1000$ & $< 3$ & $< 0.4$ & $< 1$ & $> 20$ \\
\bottomrule
\end{tabular}
}
\end{table}

\section{Details of Real-World and Simulated Experiments}
\label{app:real-robot-exp-details}

\subsection{Task Setup}
\label{app:task-setup}

\begin{figure*}[h]
    \centering
    \includegraphics[width=\textwidth]{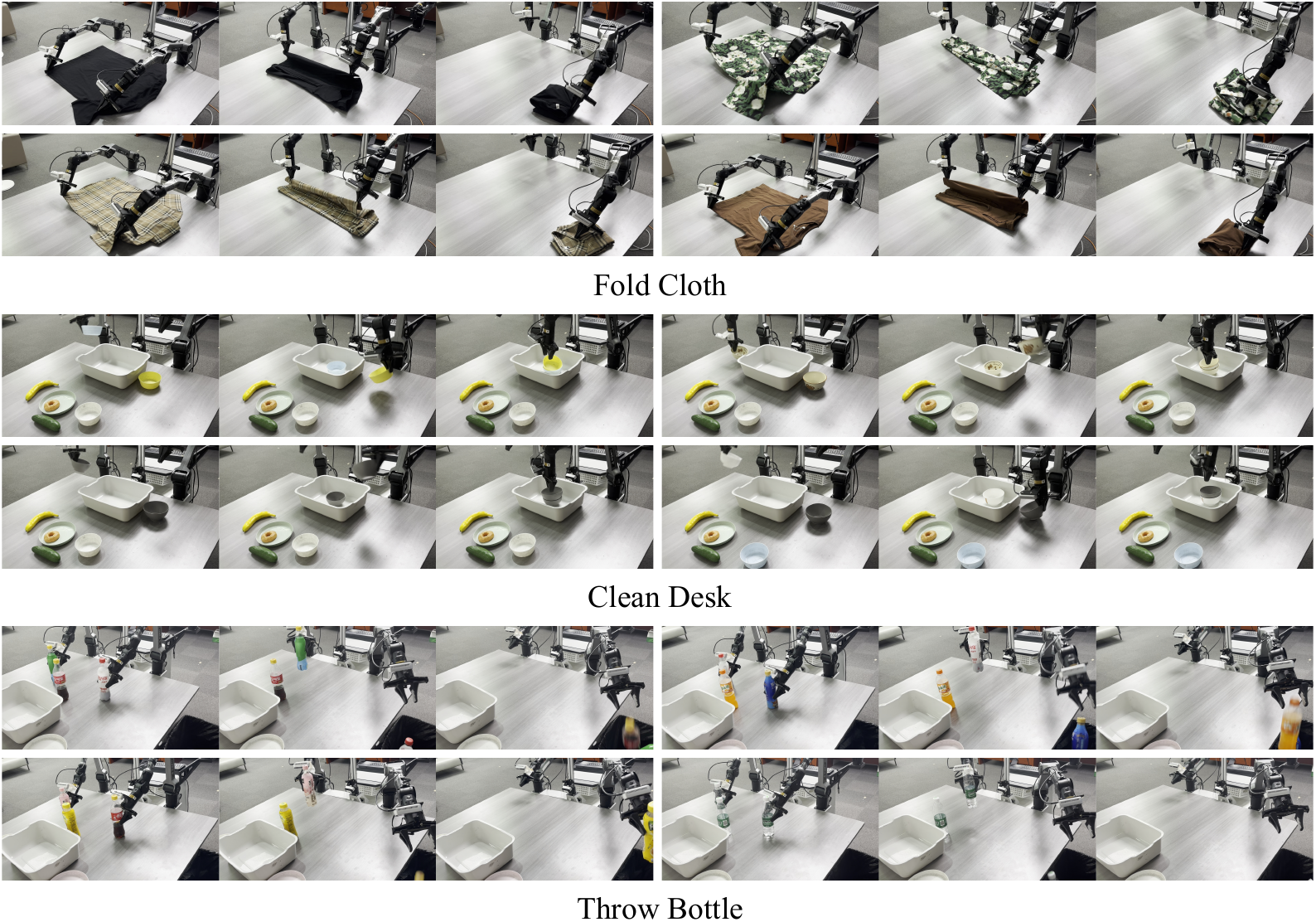}
    \caption{Real-world robot execution of $\pi_0$ policies trained with \textit{AdaMix} on three manipulation tasks.}
    \label{fig:real-eval-env}
\end{figure*}

\begin{figure*}[h]
    \centering
    \includegraphics[width=\textwidth]{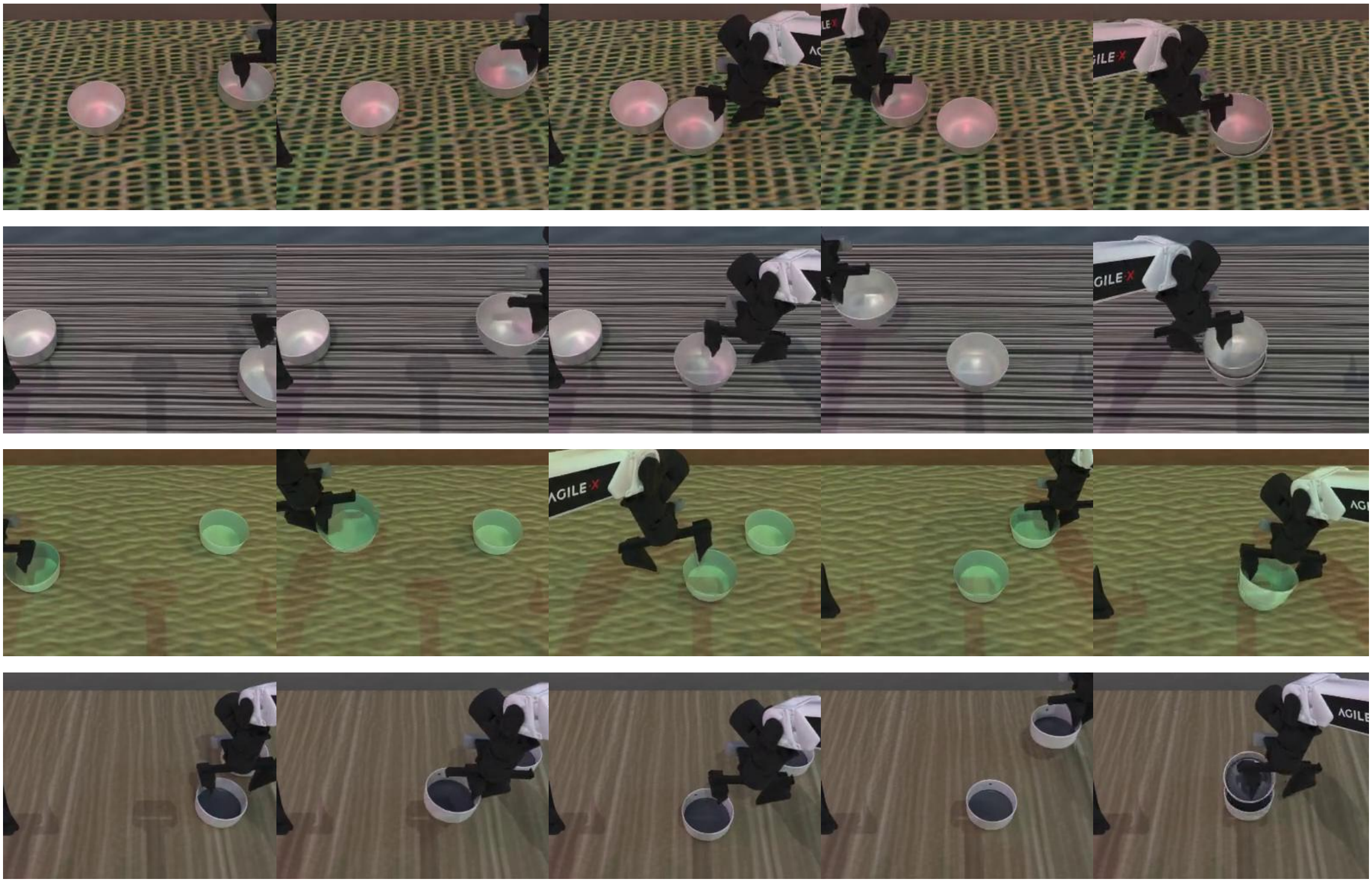}
    \caption{Simulated robot execution in RoboTwin~2.0 of $\pi_0$ policies trained with \textit{AdaMix} on the \texttt{Stack Bowls Two} task.}
    \label{fig:sim-eval-env}
\end{figure*}

Figure~\ref{fig:real-eval-env} and Figure~\ref{fig:sim-eval-env} show the execution process of each task.

\noindent
\textbf{Real-world experiments.}
We conducted experiments on the following three real-world robotic tasks, covering real-to-real and sim-to-real.

\vspace{1em}
\texttt{Fold Clothes} (Real-to-Real): This is a long-horizon, multi-stage task that involves manipulating a deformable object. 
In the first phase, the arms cooperatively fold the cloth. In the second phase, the left arm pushes the folded cloth to a designated target location on the table.

\vspace{1em}
\texttt{Clean Desk} (Sim-to-Real): In this task, a box and two bowls are placed on a tabletop. The objective is for the two robot arms to collaboratively pick up both bowls and place them inside the box.

\vspace{1em}
\texttt{Throw Bottle} (Sim-to-Real): In this task, three bottles are arranged on the table, and a trash bin is positioned adjacent to it. The robot must sequentially grasp each bottle and throw it into the trash bin.

\vspace{1em}
\noindent
\textbf{Simulation experiments.} We conducted simulation experiments in the Robo\-Twin~2.0~\cite{mu2025robotwin} on the \texttt{Stack Bowls Two}. 

\vspace{1em}
\texttt{Stack Bowls Two}:
In this task, two bowls are placed on a tabletop. The objective is for the two robot arms to collaboratively pick up both bowls and stack them at the center of the table.

\subsection{Details of Dataset and Training Configurations for VLA Policy Training}
\label{app:vla-training-config}

\begin{figure*}[h]
    \centering
    \includegraphics[width=\textwidth]{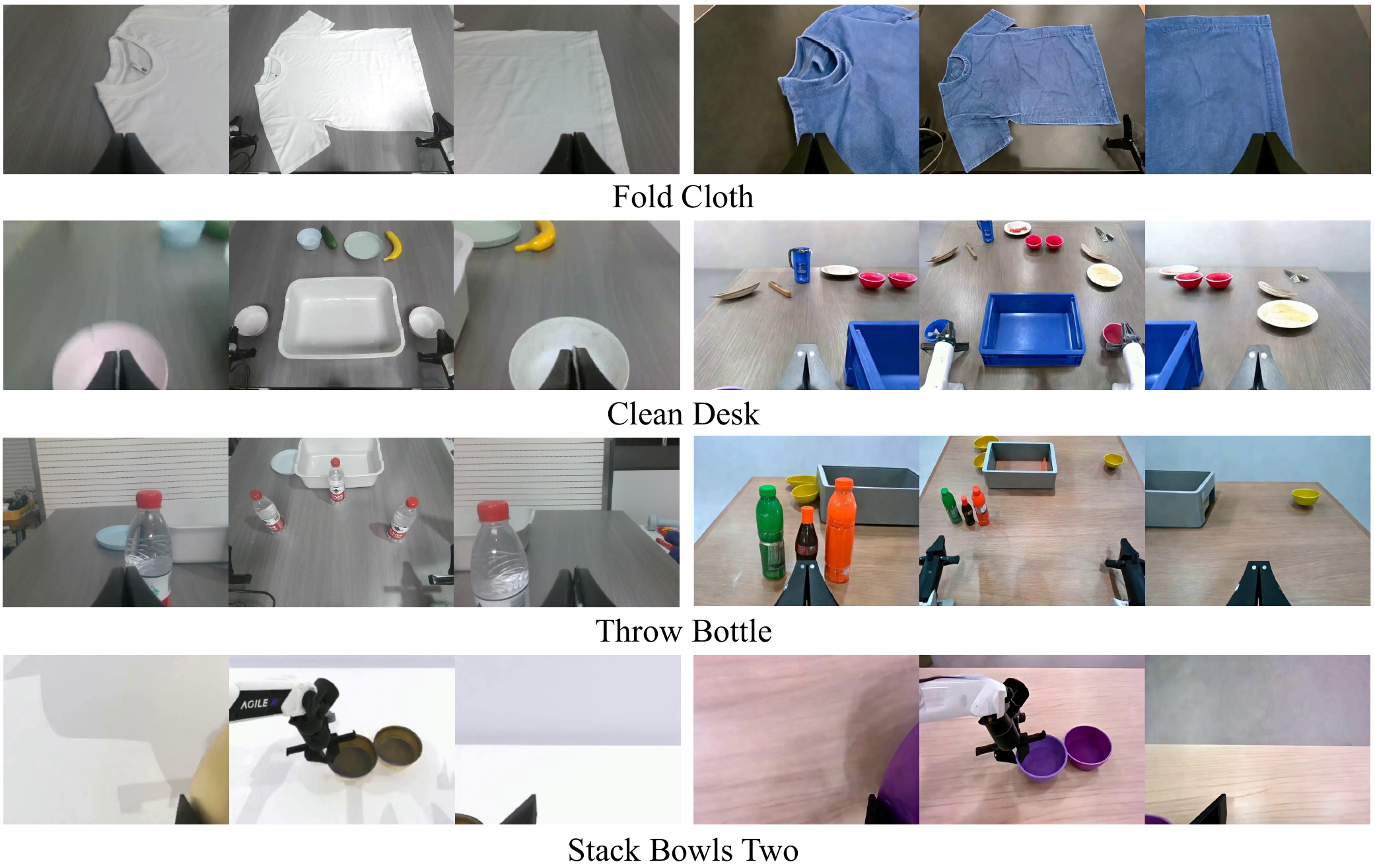}
    \caption{Samples for training the VLA policy. The left column shows real data with fixed environments and appearances (simulation data collected from simulation environments for the \texttt{Stack Bowls Two}), while the right column shows generated data with diverse object appearances.}
    \label{fig:vla-dataset}
\end{figure*}

\begin{table*}[h]
\centering
\caption{The number of real/simulation and generated data for each task used in VLA policy training, as well as the detailed training configurations on each task.
All pre-trained policies (e.g., $\pi_0$ and $\pi_{0.5}$) are fine-tuned on a single task based on their open-source pre-trained weights.}
\label{tab:vla-training-config}
\resizebox{\linewidth}{!}{%
\begin{tabular}{lcccc}
\toprule
\textbf{Configuration} & \textbf{Fold Clothes} & \textbf{Clean Desk} & \textbf{Throw Bottle} & \textbf{Stack Bowls Two} \\
\midrule
\# Real Data & 50 & 20 & 20 & -- \\
\# Simulated Data & -- & -- & -- & 50 \\
\# Generated Data & 50 & 20 & 20 & 50 \\
\# Filtered Generated Data & 46 & 19 & 19 & 47 \\
Batch Size & 64 & 64 & 64 & 64 \\
Total Training Steps ($\pi_0$) & 20000 & 5000 & 10000 & 1000 \\
Total Training Steps ($\pi_{0.5}$) & 10000 & 2500 & 5000 & 500 \\
Optimizer & AdamW & AdamW & AdamW & AdamW \\
Warmup Steps & 1000 & 1000 & 1000 & 1000\\
Init Learning Rate & $2.5 \times 10^{-8}$ & $2.5 \times 10^{-8}$ & $2.5 \times 10^{-8}$ & $2.5 \times 10^{-8}$ \\
Learning Rate Schedule & Cosine Decay & Cosine Decay & Cosine Decay & Cosine Decay \\
Trainable Parameters & All & All & All & All \\
\bottomrule
\end{tabular}
}
\end{table*}

The dataset used to train the VLA policy consists of real data and generated data, with a special note for the \texttt{Stack Bowls Two} task: its training dataset is composed of simulation data and generated data.
Real data for \texttt{Fold Clothes}, \texttt{Clean Desk}, and \texttt{Throw Bottle} is collected on the Agilex CobotMagic platform, while simulation data for \texttt{Stack Bowls Two} is collected in RoboTwin~2.0~\cite{mu2025robotwin} using the Aloha-AgileX embodiment.
For each of these tasks, the real data is collected in a single environment with a fixed set of objects, as shown in the left column of Figure~\ref{fig:vla-dataset}.
For instance, the real-world data for the \texttt{Fold Clothes} task only contains a white shirt being folded.
Generated data is synthesized using video generation models such as \textit{DreamTransfer}, as illustrated in the right column of Figure~\ref{fig:vla-dataset}.
For \texttt{Fold Clothes}, we generate appearance-diverse data in a 1:1 manner with distinct prompts for each real data.
For \texttt{Clean Desk}, \texttt{Throw Bottle}, and \texttt{Stack Bowls Two}, we similarly generate appearance-diverse data in a 1:1 ratio with different prompts for each simulation data.

This setup verifies that our model can effectively leverage generated data to improve the appearance generalization ability of the VLA policy.
The number of real/simulation and generated samples for each task used in VLA policy training, as well as the detailed training configurations, are summarized in Table~\ref{tab:vla-training-config}. 
During the experiments, we found that with the same other training configuration parameters, $\pi_{0.5}$ only requires fewer training steps to achieve performance comparable to that of $\pi_{0}$. Therefore, the number of training steps for $\pi_{0.5}$ is set to half of that for $\pi_{0}$.

\subsection{Evaluation Details}
\label{app:evaluation-details}

\subsubsection{Evaluation Environment}
Figure~\ref{fig:real-eval-env} and Figure~\ref{fig:sim-eval-env} show the real-world and simulation evaluation environments, respectively.
In the real-world environment, the foreground objects used for evaluation have \textbf{unseen} appearances that do not appear in the training set (including real data and generated data).
In the simulation environment, both the foreground objects and background exhibit \textbf{unseen} appearances that are not present in the training set (including simulation data and generated data).

\subsubsection{Evaluation Metrics Details}
In real-world and simulation experiments, we recorded all action sequences of the robot during each task execution. Each action sequence $a$ consists of $T$ actions, where the action at each time step $t$ ($t \in \{1, 2, \dots, T\}$) is a $D$-dimensional vector, corresponding to the $D$ joint degrees of freedom of the robot (in this study, $D = 14$). We comprehensively evaluate the quality of the robot's task execution based on five metrics: Success Rate, Behavior Score, Execution Time, Trajectory Smoothness, and Joint Overlimit.

\noindent
\textbf{Success rate.}
The success rate is a primary metric in traditional robot policy learning evaluation, serving as a binary indicator that reflects the proportion of successful task completions. For $N$ independent task executions, let $s_k$ be a binary variable where $s_k = 1$ if the $k$-th execution succeeds (i.e., the task goal is fully achieved) and $s_k = 0$ if it fails. The success rate is computed as:
\begin{equation}
    \text{Success Rate} = \frac{1}{N} \sum_{k=1}^{N} s_k,
    \label{eq:eq-14}
\end{equation}
where the value ranges between $0$ (all failures) and $1$ (all successes). Success is defined by the robot meeting the predefined task objectives without critical errors.

\noindent
\textbf{Behavior score.}
The behavior score is a fine-grained metric that measures task progress. It provides a more nuanced evaluation than binary success/failure by allowing partial credit. 
Specifically, we decompose each task into several critical stages. One point is awarded for the successful completion of each stage. To ensure comparability, we normalize these scores to $[0, 5]$ using the following formula:

\begin{equation}
\text{Behavior Score} = \frac{N_{completed}}{N_{total}} \times 5
\end{equation}

where $N_{completed}$ denotes the number of completed critical stages by the robot, and $N_{total}$ represents the total number of predefined critical stages for the target task.
The critical stages for each task are specified as follows:

\vspace{1em}
\noindent
\texttt{Fold Clothes}: \\
\hspace{12pt} 1. Grasp and fold the left side of the cloth. \\
\hspace{12pt} 2. Grasp and fold the right side of the cloth. \\
\hspace{12pt} 3. Grasp and fold the bottom side of the cloth. \\
\hspace{12pt} 4. Grasp and fold the top side of the cloth. \\
\hspace{12pt} 5. Push the cloth to the target position.

\vspace{1em}
\noindent
\texttt{Clean Desk}: \\
\hspace{12pt} 1. Grasp the first bowl. \\
\hspace{12pt} 2. Place the first bowl into the basket. \\
\hspace{12pt} 3. Grasp the second bowl. \\
\hspace{12pt} 4. Place the second bowl into the basket.

\vspace{1em}
\noindent
\texttt{Throw Bottle}: \\
\hspace{12pt} 1. Grasp the first bottle. \\
\hspace{12pt} 2. Throw the first bottle into the trash bin. \\
\hspace{12pt} 3. Grasp the second bottle. \\
\hspace{12pt} 4. Throw the second bottle into the trash bin. \\
\hspace{12pt} 5. Grasp the third bottle. \\
\hspace{12pt} 6. Throw the third bottle into the trash bin.

\vspace{1em}
\noindent
\texttt{Stack Bowls Two}: \\
\hspace{12pt} 1. Grasp the first bowl. \\
\hspace{12pt} 2. Place the first bowl at the center of the table. \\
\hspace{12pt} 3. Grasp the second bowl. \\
\hspace{12pt} 4. Stack the second bowl onto the first one.

\vspace{1em}
\noindent
\textbf{Execution time.}
This metric measures the time (in seconds) taken by the robot to complete one full task. Timing starts when the robot executes its first action and ends when the task is successfully completed. To ensure reliable and fair evaluation across all trials, we define specific termination criteria for each task. For the \texttt{Fold Clothes} task, completion is defined as pushing the cloth to the target location. The \texttt{Clean Desk} task is considered complete once the second bowl is placed into the basket. Similarly, the \texttt{Throw Bottle} task concludes when the last bottle is successfully thrown into the trash bin. The \texttt{Stack Bowls Two} task is completed when the second bowl is stacked onto the first one.
The average execution time is calculated only over the trials that were successfully completed.

\noindent
\textbf{Trajectory smoothness.}
We quantify trajectory smoothness using the mean of second-order joint angle differences, defined as:
\begin{equation}
    \text{Smoothness} = \frac{1}{D(T-2)} \sum_{i=1}^{T-2} \sum_{j=1}^{D} \left| \frac{\Delta^2 a_{i,j}}{(\Delta t)^2} \right|,
    \label{eq:eq-16}
\end{equation}
where $D$ denotes the total number of joint dimensions (in this study, $D = 14$), $\Delta^2 a_{i,j} = a_{i+2,j} - 2a_{i+1,j} + a_{i,j}$ represents the second-order difference of the $j$-th joint angle at time step $i$, $j \in \{1, \dots, D\}$ indexes the joints, $T$ is the total number of actions in the sequence, and $\Delta t = 1/30\ \text{s}$. The term $T-2$ accounts for the valid range of $i$ when computing second-order differences. 

\noindent
\textbf{Joint overlimit.}
The Joint Overlimit metric quantifies the number of actions in the sequence where at least one joint angle exceeds its predefined limit. Formally, let $\Omega_j = [\underline{\theta}_j, \overline{\theta}_j]$ denote the valid angle range for the $j$-th joint ($j \in \{1, 2, \dots, D\}$), where $\underline{\theta}_j$ and $\overline{\theta}_j$ are the lower and upper limits of the $j$-th joint, respectively. For each action $a_i = [a_{i,1}, a_{i,2}, \dots, a_{i,D}]$ at time step $i$ ($i \in \{1, 2, \dots, T\}$), we define an indicator function:
\begin{equation}
    \mathbb{I}(a_i) = 
    \begin{cases} 
    1 & \text{if } \exists j \in \{1, 2, \dots, D\} \text{ such that } a_{i,j} \notin \Omega_j, \\
    0 & \text{otherwise}.
    \end{cases}
\end{equation}
The Joint Overlimit metric is then expressed as:
\begin{equation}
    \text{Overlimit} = \sum_{i=1}^{T} \mathbb{I}(a_i),
    \label{eq:eq-18}
\end{equation}
where $T$ is the total number of actions in the sequence, and $a_{i,j}$ represents the angle of the $j$-th joint in the $i$-th action.

\section{Qualitative Comparison with Other Video Generation Models}
\label{app:video-generation-models-comparison}

\begin{wrapfigure}{r}{6cm}
    \vspace{-2em}
    \centering 
    \includegraphics[width=\linewidth]{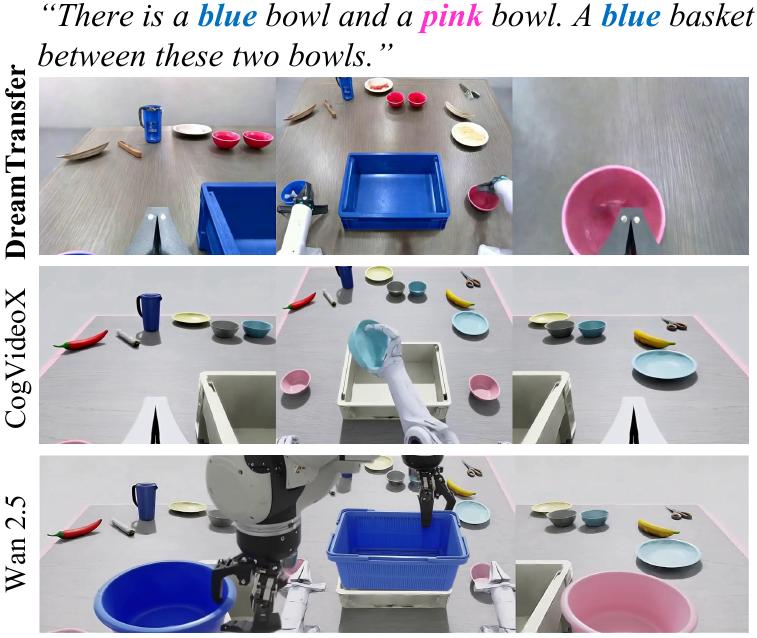}
    \caption{Qualitative comparison between \textit{DreamTransfer} and general video generation models (CogVideoX~\cite{yang2025cogvideox} and Wan2.5~\cite{wan2025wan}) in embodied scenarios.}
    \vspace{-1em}
    \label{fig:general-video-gen-comparison}
\end{wrapfigure}

While general video generation models exhibit strong open-domain performance, Figure~\ref{fig:general-video-gen-comparison} reveals their limitations in embodied scenarios, specifically multi-view inconsistency and poor physical fidelity. Robotic task-specialized models such as Cosmos-Transfer1~\cite{nvidia2025cosmostransfer} and RoboTransfer~\cite{liu2025robotransfer} leverage large-scale embodied data to achieve better kinematic plausibility, and we compare \textit{DreamTransfer} against these specialized models.

Figure~\ref{fig:transfer-compare-demo} presents a qualitative comparison between \textit{DreamTransfer} and existing embodied manipulation video transfer methods~\cite{liu2025robotransfer,nvidia2025cosmostransfer}. Our approach generates videos that maintain consistent geometry across multiple views, preserve accurate 3D structure, and respond faithfully to text-driven appearance edits, significantly outperforming prior models in visual plausibility and controllability.

\begin{figure*}[h]
    \centering
    \includegraphics[width=\textwidth]{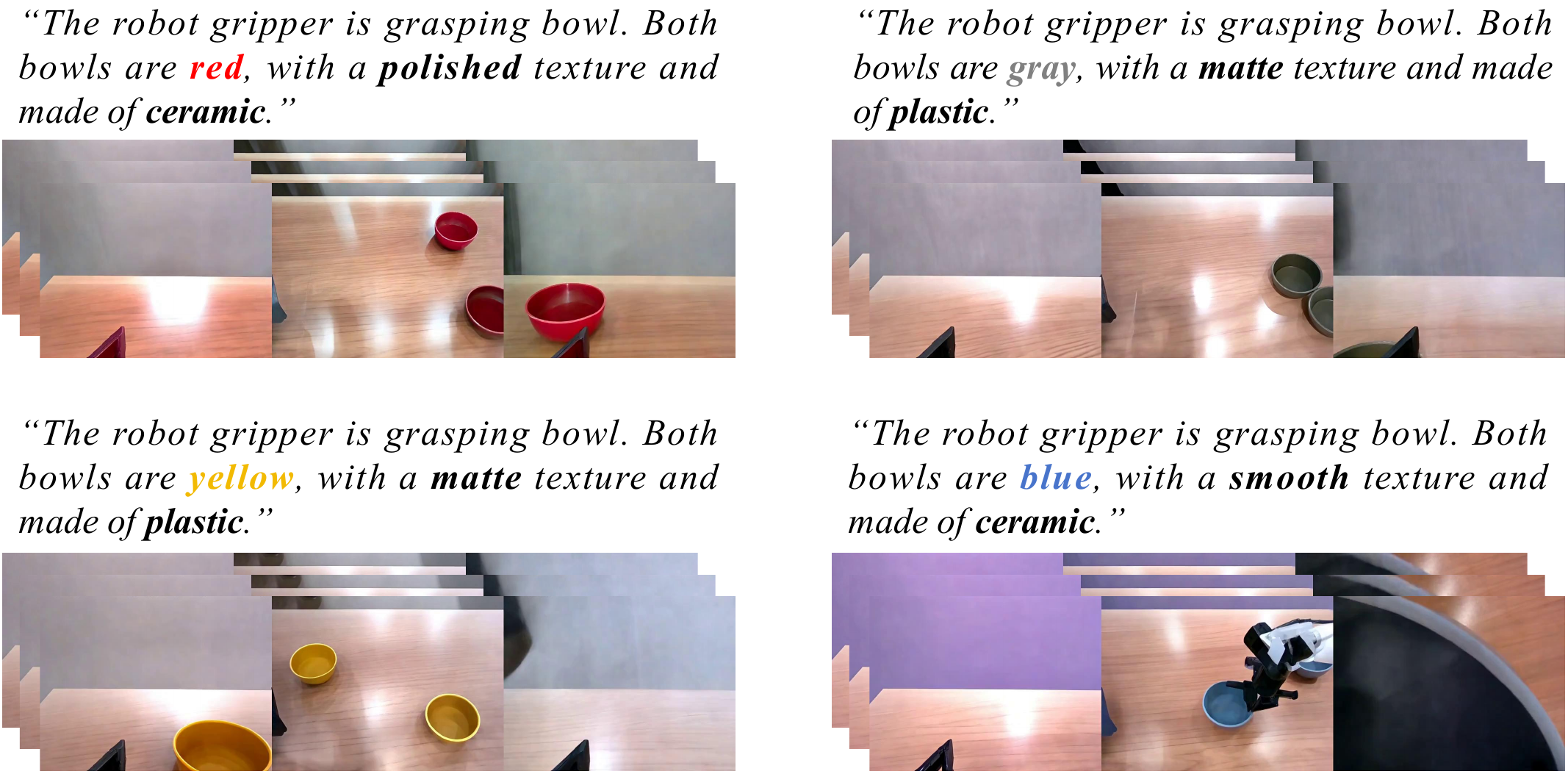}
    \caption{Diverse data generated by \textit{DreamTransfer} from the simulation data (first column of the last row in Figure~\ref{fig:vla-dataset}) for the \texttt{Stack Bowls Two} task. \textit{DreamTransfer} exhibits strong controllability in embodied manipulation video generation.}
    \label{fig:simulation-transfer}
\end{figure*}

Figure~\ref{fig:simulation-transfer} shows diverse data generated by \textit{DreamTransfer} from the single-appearance simulation data (first column of the last row in Figure~\ref{fig:vla-dataset}) for the \texttt{Stack Bowls Two} task.

\section{Qualitative Results of Policy Deployment}
\label{app:robot-demo}

We evaluated the VLA policy trained with \textit{AdaMix} in both real-world and simulation environments. 
In the real-world environment, we deployed the policy on the Agilex CobotMagic platform with two PiPER arms to execute the \texttt{Fold Clothes}, \texttt{Clean Desk}, and \texttt{Throw Bottle} tasks.
In the RoboTwin~2.0~\cite{mu2025robotwin} environment, we deployed the policy on the Aloha-AgileX embodiment to perform the \texttt{Stack Bowls Two} task.
Figure~\ref{fig:real-eval-env} and Figure~\ref{fig:sim-eval-env} present the deployment results of the $\pi_0$ policy trained with \textit{AdaMix} in real-world and simulation environments, respectively.

These policies demonstrate the ability to generalize to real-world and simulation scenes containing objects with appearances unseen during training.